\documentclass{interact}
\usepackage{cite}
\usepackage{amsmath,amssymb,amsfonts}
\usepackage{algorithm}
\usepackage{algpseudocode}
\usepackage{graphicx}
\usepackage{textcomp}
\usepackage{xcolor}

\usepackage{caption} 
\usepackage{multirow}
\usepackage{hhline}
\usepackage{color,soul} 

\def\BibTeX{{\rm B\kern-.05em{\sc i\kern-.025em b}\kern-.08em
    T\kern-.1667em\lower.7ex\hbox{E}\kern-.125emX}}
\begin{document}
\title{Autonomous Industrial Assembly using Force, Torque, and RGB-D sensing
}

\author{
\name{James Watson\textsuperscript{a}\thanks{james.watson-2@colorado.edu}, Austin Miller\textsuperscript{b} and Nikolaus Correll\textsuperscript{a,b}}
\affil{\textsuperscript{a}Department of Computer Science, University of Colorado, Boulder, CO 80309, USA \\
\textsuperscript{b}Robotic Materials Inc., Boulder, CO 80301, USA}
}

\maketitle
\begin{abstract}
We present algorithms and results for a robotic manipulation system that was designed to be easily programmable and adaptable to various tasks common to industrial setting, which is inspired by the Industrial Assembly Challenge at the 2018 World Robotics Summit in Tokyo.  
This challenge included assembly of standard, commercially available industrial parts into 2D and 3D assemblies. We demonstrate three tasks that can be classified into ``peg-in-hole'' and ``hole-on-peg'' tasks and identify two canonical algorithms: spiral-based search and tilting insertion. Both algorithms use hand-coded thresholds in the force and torque domains to detect critical points in the assembly.  
After briefly summarizing the state of the art in research, we describe the strategy and approach utilized by the tested system, how it's design bears on its performance, statistics on 20 experimental trials for each task, lessons learned during the development of the system, and open research challenges that still remain.
\end{abstract}


\textit{Preprint, Approved for publication in Advanced Robotics: Received 16 Jul 2019, Accepted 12 Nov 2019, Published online: 01 Feb 2020}

\section{Introduction} 
The Industrial Assembly Challenge at the 2018 World Robotics Summit in Tokyo was designed to be a test of the reliability and  flexibility of automated manufacturing systems.  The challenge was comprised of three exercises common to industrial settings; kitting --- picking a collection of individual parts from bins and placing them in a tray in preparation for a mock final assembly, Task Board --- individual assembly tasks separated from each other on a 2D plane, and 3D assembly --- a complex mock product that requires multiple part reorientations and multiple insertion directions. 


An end-to-end automated assembly system must be able to identify, and recover from, task failures in order to be considered truly automated \cite{FDICNN}.  This work represents an initial step in this direction.  We present a robotic system comprised of a serial manipulator, with vision and force sensing capability.  We then benchmark the performance of algorithms for several common industrial assembly tasks.  We qualitatively evaluate the force data from successful and failed task attempts using the algorithms and comment on events that occurred during task execution that precipitated failure.

For this work we focus on the second exercise, the Task Board, because it provides an ideal test environment for robust assembly algorithms in a constrained but realistic setting.  The task board is comprised of components common to manufactured products, available from Misumi.  Each of the actions are performed in isolation from the others.

\begin{figure}[!htb]
	\centering
	\includegraphics[width=0.75\columnwidth]{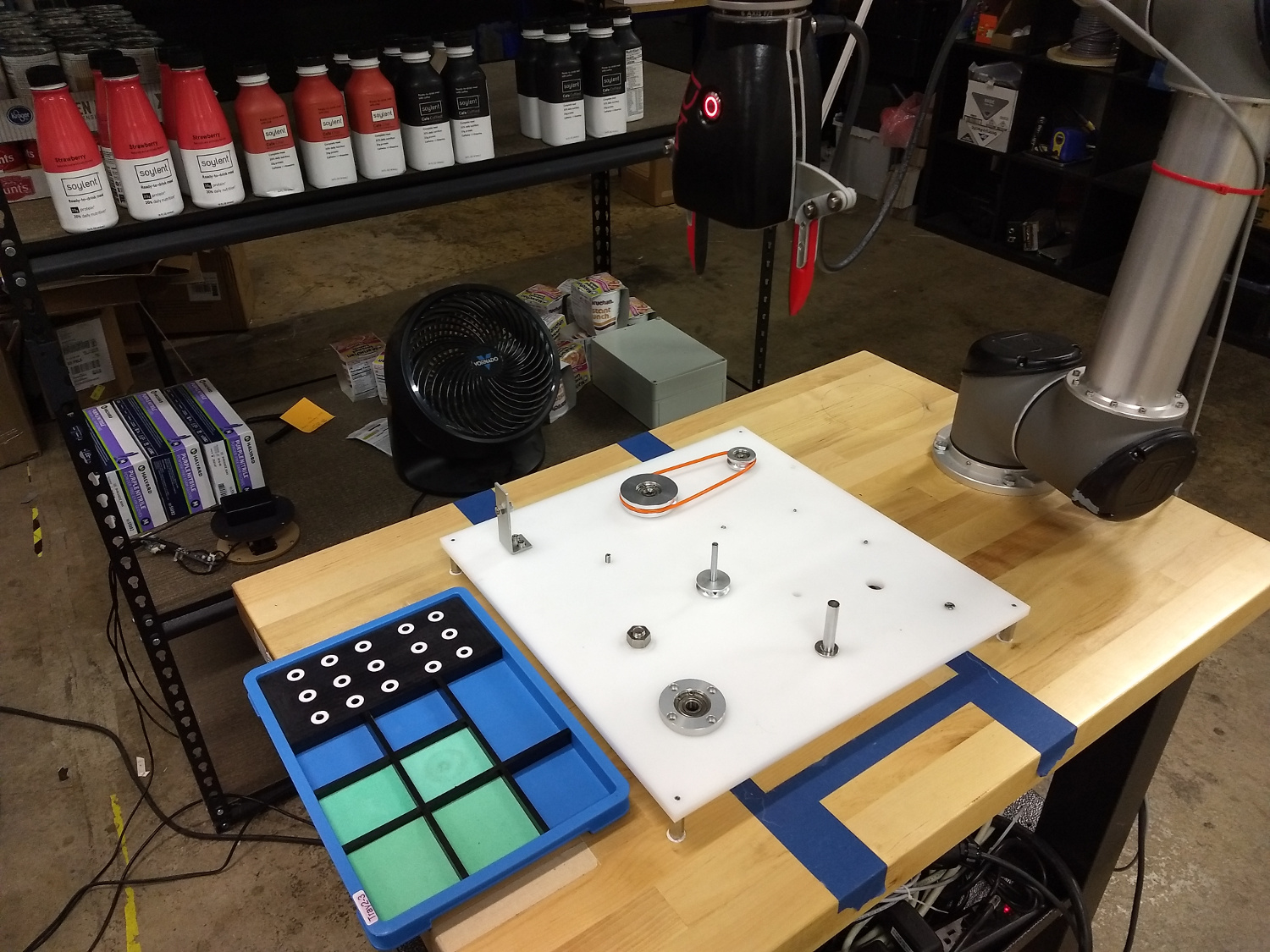}
	\caption{Experimental Setup. All parts are presented in a ``kitting tray'' (left) and need to be assembled on a plastic plate (right).}
	\label{photoAfter}
\end{figure}

\subsection{Contribution of this paper}
In this paper, we perform a detailed experimental study on three selected tasks from the assembly challenge that are representative of ``peg-in-hole'' and ``hole-on-peg'' assemblies, for which we present two strategies, spiral-based search and tilt-based insertion, each of which applies to both cases. We demonstrate how these primitives can be combined with 3D perception to identify part, hole and pegs, and other force-based primitives to perform complex, multi-step assemblies. We record both torque and force measurements over time for both successful and failed assemblies, which are available online\footnote{https://github.com/RoboticMaterials/wrs2018-paper}, and discuss future directions of research that will help increase performance of autonomous assembly of industrial products.   

\section{Previous Work} 
Autonomous robotic assembly is distinct from conventional industrial assembly in that part and placement locations are subject to uncertainty. There exist a rich body of work for autonomous assembly of truss structures in space \cite{dorsey2012efficient,belvin2016space}. For example, Komendera \textit{et al.} \cite{komendera2015precise} presents a method for sensor-based construction of complex truss structures. Space is a challenging domain for autonomous assembly as the cost of humans in space far outweigh potential savings from increased throughput, which is paramount in the industrial domain. 

The system presented in this work is based on the composition of assembly task strategies from smaller parametric tasks. N{\"a}gele \textit{et al.} \cite{prototypeCompose} have proposed such a system, and define a framework for how skills might be composed.  Their aim is to avoid the repeated work of developing and tuning similar actions by using previously-deployed actions as prototypes, then incrementally refining the parameters of the cloned actions.  The proposed system has a high potential for flexibility, but requires a complex framework to model the interactions between skills, controllers, and movement constraints.  Halt \textit{et al.} \cite{asmPrimitives} identify several primitive actions that can be composed into larger assembly tasks.  Each primitive action has feature coordinates that are mapped to task space coordinates, which impose constraints on motion and how primitives may be combined.  They build on the iTASC \cite{iTASC} framework, which attempts to fulfill multiple objectives with plans composed of these primitives.  However, as the authors state, the automated sequencing and combination of primitive actions based on their constraints and transition conditions remains an open area of research.

We also present a qualitative analysis of the force and torque signatures of our assembly tasks, and interpret their physical significance as it relates to task success or failure.  Much work has been done to automate failure detection, and this is a necessary step in the development of fully automated assembly systems. 
Lopes \textit{et al.} \cite{lopes1998feature} present a series of hand-coded features to classify force-torque time series for various robot grasping scenarios that are relevant during robotic assembly \cite{camarinha1996integration}. Failures can be identified using a support vector machine classifier with only 65 training examples using a sensor with a single axis only, as demonstrated by Rodriguez \textit{et al.} \cite{forceTraceSVM}.  Once trained, the classifier only applies to one task, and classification can only take place after the action has completed.  Expert knowledge about task progression was encoded into a state machine by Majdzik \textit{et al.} \cite{FSMclassify}.  Their system can identify multiple failure modes, but it relies on a series of complex rules to achieve this feat.  Perhaps the most important capability of a flexible assembly system is the capacity to identify failures in-progress, so that costly mistakes and wasted time may be prevented.  Moreira \textit{et al.} \cite{FDICNN} presented an online, failure-detecting convolutional neural network classifier that can predict failure of threading a nut on a stud during action execution.  Their method had similar efficacy to multi-layer perceptron and support vector machine methods applied to the same task. 

We add to this body of work by providing detailed time-series data of a series of realistic assemblies that are fully perception based, i.e. in which both initial part and assembly location are subject to uncertainty and are detected by the robot using RGB-D perception. 

\section{Materials and Methods} 

\subsection{Task Description} \label{taskDesc}

The setting for the tasks is a base plate 1 cm thick acetal resin that has holes for mounting the free components used on the 2D assembly task.  The plate is supported by four standoffs which are each 3 cm high. There are three tasks in the experiment: Large Bearing Task, Small Pulley Task, and Stud Task, which are taken from the Taskboard Challenge at the World Robotics Summit.

The goal of the large bearing task is to lower a flanged bearing with housing (Misumi part SBARB6200ZZ-30) into a hole in the base plate.  The hole in the base ($\oslash 36+0.2mm$) is a sliding fit with the large bearing, such that an insertion attempt that is out of alignment will easily jam the bearing and prevent insertion. This task is representative for a ``peg-in-hole'' task. 

For the small pulley task, a pulley wheel 3.2 cm in diameter (Misumi part MBRFA30-2-P6) is placed onto a 6 mm diameter shaft that is 5.2 cm long (Misumi part PSSFAN6-50-F10-B8-P4). This task is representative for a ``hole-on-peg'' task. 

In order to accomplish the large nut task, the nut (Misumi part SLBNR12) must be threaded onto a stud (Misumi parts SCB12-25, screw, and SPWF12, washer) in the base plate. This task is a combination of a ``hole-on-peg'' task combined with a screwing task. 


\subsection{Robotic System}

Control of the system is provided by a Robotic Materials \emph{SmartHand}.  The SmartHand is an integrated computing, vision, and parallel gripper platform designed for use with serial, collaborative \cite{collabSafety} robots.  It integrates an nVidia Jetson TX2 computer for control and image processing.  Visual sensory data comes from an Intel RealSense D430 that is mounted in the palm.  The RealSense is a structured-light camera that projects two infrared patterns into the field of view of the image sensor, and calculates per-pixel depth using interferometry, resulting into a Red-Green-Blue plus Depth image (RGB-D). Both RGB-D and infrared images can be obtained from the sensor.  The camera and computer are mounted to an internal frame that provides passive cooling.  Smarthand interacts with its environment through a parallel gripper.  The fingers of the gripper have a tapered profile that mimics the beak of a crow, a design choice based on the renown of corvids for dextrous manipulation \cite{corvidAssembly}.  Also like a crow, the vision sensor's field of view is in the immediate workspace of the gripper, allowing the gripper to see assembly objects as close as 11cm.

The SmartHand is mounted to an OptoForce (now OnRobot) HEX-E 6-axis force-torque (FT) sensor capable of sensing the wrench at the wrist joint with a resolution of 0.2 Newtons in X-Y and 0.5 Newtons in Z direction for force measurements and 0.010 in X-Y and 0.002 Newton-meters around the Z direction for torque measurements.

The FT sensor is mounted to a Universal Robotics UR5 serial manipulator.  The UR5 employs sensitive motor torque feedback to a safety system that prevents it from imparting harmful forces to humans and objects in its workcell.  Thus, UR5 is considered to belong to a class of machines known as collaborative robots \cite{cobot,UR5manual}, which have been steadily gaining popularity in both industry \cite{IndustryCollabIncrease} and research \cite{EduCollabIncrease}.

\section{System Description}

First, we describe the design and configuration of the system used to execute our experiments.  Then, we describe the control software used for the system and how it relates to advantageous modularization of common assembly tasks.

\subsection{Robotic System and Apparatus}

The control software used for this work is an extension of the RMStudio software, which is a collection of Python 3 libraries for interface with the UR5 robot and SmartHand via Jupyter Notebooks. Here, all sensors and actuators are abstracted into Python libraries that communicate with the sensor and actuator programming interfaces using XML-RPC, RTDE and socket communications, resulting into a single Python script that controls all operations. The Jupyter notebook allows the programmer to execute code step-by-step, visualizing output inline with code, which supports rapid prototyping.

Parts and holes are located using appropriate methods from the \emph{Open3D} and \emph{OpenCV} libraries. 

To move the robot, we make use of UR's inverse kinematics interface, which allows us to move the robot in Cartesian space. Here, we avoid collisions by moving the robots to safe way-points above the task board, which we assume to remain unobstructed during task progress. 

\subsection{Task Modularization}

Modularization and robustness are our main goals in the design of assembly actions.  By breaking up tasks into common actions, we are able to reuse actions for multiple tasks. Halt \textit{et al.} \cite{asmPrimitives} identify 6 primitive actions in the execution of assembly tasks.  We have designed the actions for this system to be robust to small misalignment and displacement, so that single-purpose jigging can be avoided as much as possible.  Presently, our system has two categories of actions that meet this need: \emph{perception steps} and \emph{join actions}.

\subsection{Perception steps}

\begin{figure}[!htb]
	\centering
	\includegraphics[width=0.45\columnwidth]{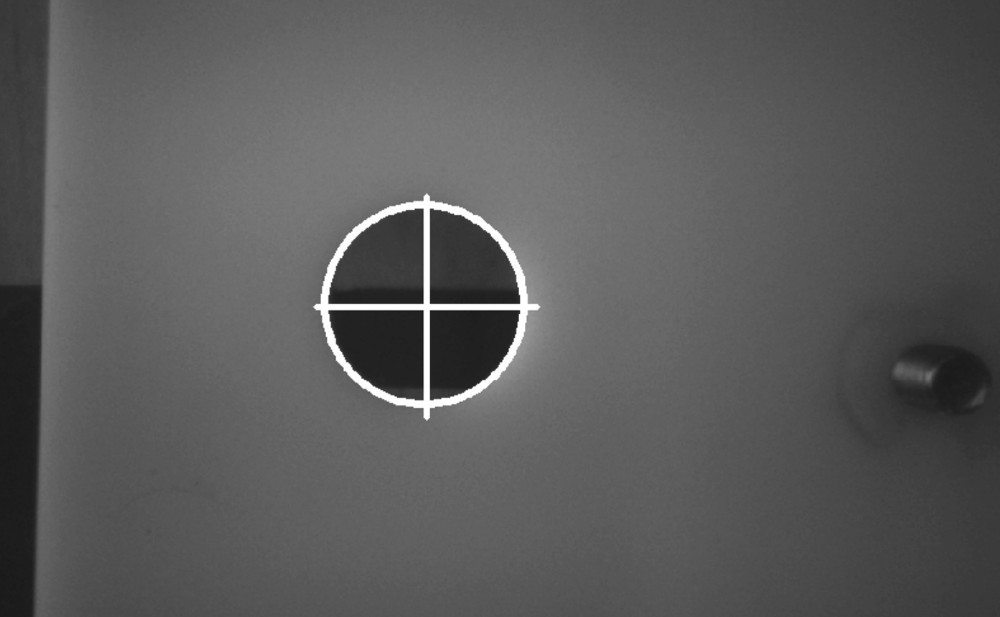}
	\includegraphics[width=0.45\columnwidth]{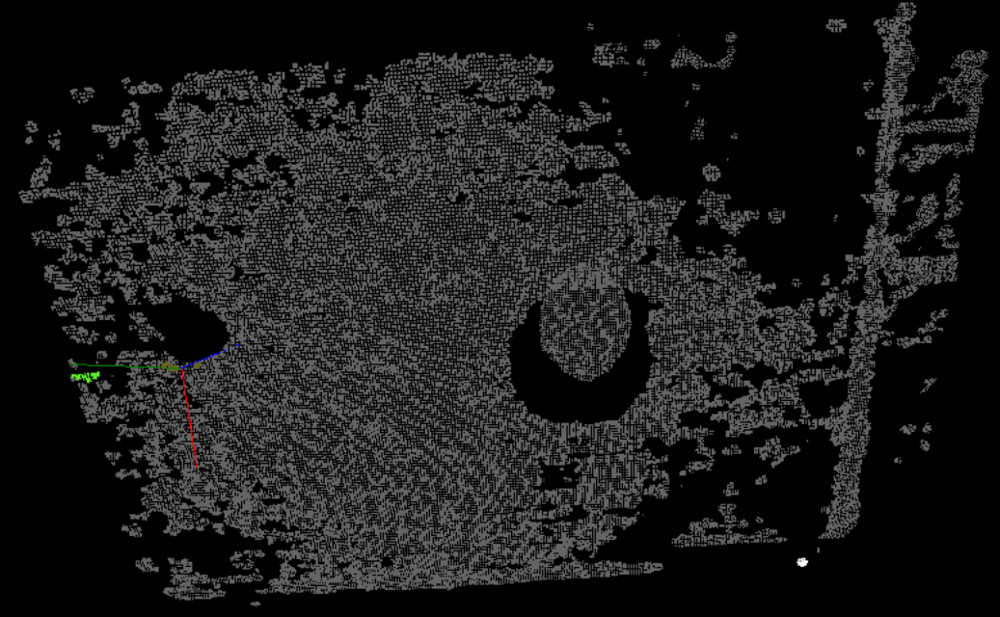}
	\caption{Left: Infrared image of the bearing hole. The center of hole has been located by \textsc{CircleLocator}, as shown by the white overlay. Point cloud of the bearing hole. Generated at the same camera position as the infrared image.}
	\label{IRcircle}\label{depthcircle}
\end{figure}

Perception steps are designed to obtain a more accurate pose estimate of a part that will participate in a later join action.  We have created two different perception steps that have wide applicability to the assembly tasks that comprise the competition assembly problems.  The first is a circular feature locator, \textsc{CircleLocator}.  This has obvious uses for fasteners, shafts, and the associated holes.  Fasteners comprised a large portion of the WRS Challenges, so this function was deployed repeatedly in our solution to the assembly problem.  The hole pose locator makes use of the Circle Hough Transform \cite{houghCircle} and camera intrinsics in order to obtain a hole pose (Figure \ref{IRcircle}, left).  The circle locator takes the diameter of the queried feature and the distance between the plane of the circle and the gripper camera as parameters.  \textsc{CircleLocator} assumes that the camera begins the procedure with the line of sight of the camera pointed in a direction perpendicular to the circle feature plane.  It uses the RealSense field of view (FOV) parameters to calculate how large the circle should appear in the captured image.  Although this has wide use, it has the drawback of fixing only 2 of the 3 SO(3) degrees of freedom.  For these cases, the point cloud feature locator provides a means for us to match a known point model with a perceived model.  This method relies on iterative closest point (ICP) \cite{ICP}, to match a user-recorded example point cloud in a desired orientation to a point cloud scanned by the hand camera.

\subsection{Part placement actions} \label{placeProcs}
The above part location strategies proved useful for providing finer locations of parts in an environment where the parts are arranged by hand.  Even so, there was still variation in how the parts were gripped by the robot.  In order to account for this, we developed part placement actions designed to be robust against these variations.  There are two actions in this category: spiral and tilt-in.  Both of these actions are suitable for peg-in-hole and hole-on-peg operations, which are the focus of this work.  

The \textbf{Pickpart} procedure is designed as a simple part grasping action.  It takes the center of the grasp ($\vec{x}_{PartLocation}$), diameter of the part at the grasp pose (Dia$_{Bearing}$), and depth from the part surface ($z_{PickDepth}$) as parameters. $z_{PickDepth}$ is expressed in the gripper's local frame, and is used to tune how much of the gripper finger surface engages the part.  Once the gripper reaches the part location and applies the offset, the gripper closes as fully as possible.  This procedure does not provide feedback as to whether the part was grasped.

Although it is not a placement action, exactly, we designed \textbf{TampWithForceLimit} as a compensatory procedure to deal with part jamming that might occur during assembly.  It simply moves the gripper to an absolute location $\vec{x}_{loc}$, then moves the gripper down in its local $-Z$ axis until either the movement limit $\Delta z$ is reached, or the wrist sensor detects a reaction force that exceeds the user-specified limit $F_{touch}$.  It then moves the gripper back along the approach vector in the opposite direction by $\Delta z$.

\textbf{Twist} is another compensatory procedure; it is meant to free a part that has mechanically bound (jammed) on its fulcrum. It is designed to rotate the bound part about its fulcrum in an oscillatory way. \textbf{Twist} takes the grasp center $\vec{x}_{loc}$, the angular oscillation magnitude $\theta_{osc}$, and the number of oscillations $N_{osc}$ as parameters.  $\vec{x}_{loc}$ must be defined so that the local gripper $Z$-axis is aligned with the fulcrum axis in the lab frame.  The procedure moves the gripper to the grasp center, closes the gripper, then sweeps the wrist $Z$-axis rotation by $\pm \theta_{osc}$ for $N_{osc}$ repetitions.

The \textbf{Thread} procedure is used for the nut threading task.  It takes the beginning nut pose $\vec{x}_{loc}$, number of flats to advance each turn $N_{flats}$, tightening torque limit $T_{z,limit}$, and fastener thread pitch in $d_{pitch}$ as parameters.  The procedure begins by closing the gripper at the grip center, assuming that $\vec{x}_{loc}$ has been defined with an orientation that places the gripper finger surfaces on flats of the nut.  Then, in a loop, the gripper rotates by $-60 N_{flats}$ degrees (clockwise) while moving in $-Z$ by the thread pitch, opens, and then rotates back by $60 N_{flats}$ degrees counterclockwise.  The loop exits when the $T_{z,limit}$ is reached.  $T_{z,limit}$ is chosen so that then nut is finger tight.  This limit should be encountered whether the nut has reached the base plate or cross-threaded.

\begin{figure}
    \centering
    \includegraphics[width=\columnwidth]{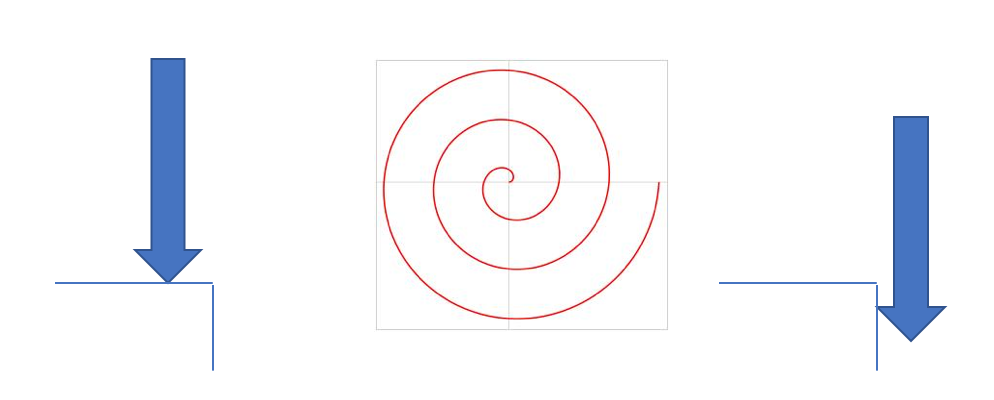}
    \caption{Key steps during hole-on-peg assembly by spiraling motion. From left to right: move to contact surface until contact, perform spiraling motion from inside out (top-view), move down when possible.}
    \label{fig:spiral}
\end{figure}

\subsection{Action Algorithms} \label{ActionAlgos}

The spiral part insertion algorithm is provided in pseudo-code in Algorithm \ref{alg:spiral} and illustrated in Figure \ref{fig:spiral}.  For the spiral action, the held part is brought towards the reference part along the axis of insertion at the perceived location of the hole until a contact force is sensed.
For this, the robot first moves to an absolute position (\textbf{MoveAbs()}) given by the insertion pose $\vec{x}_i$ plus an offset $\Delta z$ above $\vec{x}_i$. Note, that $\vec{x}_i$ is a 6-DoF pose, which encodes not only a 3D point, but also the pose of the insertion plane. The robot then moves towards $\vec{x}_i$ along the insertion axis until it has either traveled $\Delta_{max}$ or the function \textsc{MaxDownForce} does not evaluate true given a certain threshold. 
This is indicated by the left arrow construct. Using a reasonable value $\Delta_{max}>\Delta z$, the robot moves with speed $v$ toward the insertion location, until a certain contact force $-F_t$ is exceeded. 
The robot then performs a spiraling motion \textbf{MoveSpiral()} with step-size $\Delta s$,  maximum radius $r_{max}$,  speed $v$ and acceleration $a$  as long as \textsc{MaxSpiralForces()} returns true. The spiral equation is given by 
\begin{align*}
    x_i &= x_0 + r_i \cos{\phi_i}  \\
    y_i &= y_0 + r_i \sin{\phi_i}  
\end{align*}
where ($x_0,y_0$) is some starting position, $r_i$ is a radius incremented every step, and $\phi_i$ is an angle updated every step.

The intent of this operation is to use the end of the held part to search the surface of the stationary part for the actual location of the hole.  When the held part reaches the hole, then the downward pressure causes the part to drop into the hole.  When the held part has entered the hole, it is confined on two translation axes, leading to excessive lateral forces and torques. Specifically, this function is written to detect excessive torques (here: 0.9 Nm) in either direction around the X and Y axes, as well as pressure forces exceeding $F_d$. Should these events occur, it is likely that the part has snapped into its final location and even small movements immediately lead to large torques or contact forces.

\begin{algorithm*}[!htb]
    \caption{Algorithm for inserting a part using a spiral movement} \label{alg:spiral}
    \begin{algorithmic}[1]
        \Procedure{InsertPartSpiral}{$\vec{x}_i$, $\Delta z$, $F_t=1$, $F_d=1$, $F_i=2$, $\Delta_{max}$}
        \State
        \Function{MaxDownForce}{f}\Comment{Used as a stopping criterion when moving down}
            \State \Return $GetWristForce(Z)< f$
        \EndFunction
        \State
        \Function{MaxSpiralForces}{}\Comment{Used as a stopping criterin during spiraling}
            \State \Return $(\|GetWristTorque(X)\| > 0.9) \lor (\|GetWristTorque(Y)\| > 0.9) \lor (GetWristForce(Z) > -F_d)$
        \EndFunction
        \State
        \State \textbf{MoveAbs} ($\vec{x}_i$ + $[0,0,\Delta z,0,0,0]$)
        \State \textbf{MoveRel} ($[0,0,-\Delta_{max},0,0,0],v=0.01$) $\leftarrow$ \textsc{MaxDownForce($-F_t$)}
        \State \textbf{MoveSpiral}($x_0$,$y_0$,$\Delta s=0.00001$,$r_{max}=0.004$,$v=0.002$,$a=0.5$) $\leftarrow$ \textsc{MaxSpiralForces}()
        \State \textbf{MoveRel} ($[0,0,-\Delta_{max},0,0,0],v=0.01$) $\leftarrow$ \textsc{MaxDownForce($-F_i$)}
        \EndProcedure
        
    \end{algorithmic}
\end{algorithm*}

\begin{figure}[!htb]
    \centering
    \includegraphics[width=\columnwidth]{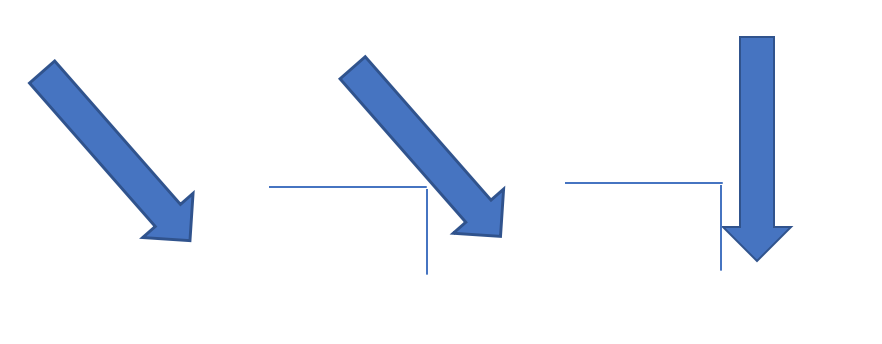}
    \caption{Schematic representation of the progression of the tilt-in action}
    \label{fig:tiltPic}
\end{figure}

The tilt-in action is also designed for cylindrical peg-in-hole operations and is provided as pseudo-code in Algorithm \ref{alg:tilt} (illustrated in Figure \ref{fig:tiltPic}). In this operation, the tip of the held part is brought above and near to the hole located at $\vec{x}_{hole}$ (with diameter Dia) in the stationary part with the hole axis and held part axes aligned. Then, the held part is tilted away from the hole axis by $\theta_{tilt}$ and translated in the direction of the tilt by $\Delta x$.  The tilted part is lowered into contact with the hole edge, using maximum normal force $F_{touch}$ as the stopping criterion to indicate contact.  If aligned properly, the arc of the lowest held part edge now dips below the level of the hole.  After contact is made, the held part is tilted $-\theta_{tilt}$ back into alignment with the hole axis with pressure applied. Then, an additional downward motion is attempted with a maximum normal force limit of $F_{insert}$ in an effort to seat the held part in its hole.

\begin{algorithm*}[!htb]
    \caption{Algorithm for inserting a part using a tilting movement} \label{alg:tilt}
    \begin{algorithmic}[1]
        \Procedure{InsertTilt}{ $\vec{x}_{hole}$ , $\Delta z$ , $\theta_{tilt}$ , Dia , $\Delta x$ , $F_{touch}$ , $F_{insert}$ , $\Delta_{max}$ }
        \State 
        \State \textbf{MoveAbs} ($\vec{x}_{hole} + [0,0,\Delta z,0,0,0]$)
        \State $x_{offset} \leftarrow \Delta x + \sin{ \theta_{tilt} }$ 
        \State \textbf{MoveRel}( $[ -( x_{offset} + \text{Dia}/4 ) , 0 , 0 , 0 , 0 , 0 ]$ )
        \State \textbf{MoveRel}( $[ 0 , 0 , 0 , 0 , \theta_{tilt} , 0 ]$ )
        \State \textbf{MoveRel} ($[0,0,-\Delta_{max},0,0,0]$) $\leftarrow$ \textsc{MaxDownForce($-F_{touch}$)}
        \State \textbf{MoveRel}( $[ ( x_{offset} + \text{Dia}/4 ) , 0 , 0 , 0 , -\theta_{tilt} , 0 ]$ ) $\leftarrow$ \textsc{MaxDownForce($-F_{insert}$)}
        \State \textbf{OpenGripper}()
        \State \textbf{MoveRel} ($[0,0,\Delta_{max},0,0,0]$)
        \EndProcedure
    \end{algorithmic}
\end{algorithm*}

\subsection{Task Strategy}

Using the action algorithms described in Section \ref{ActionAlgos}, we composed task-level algorithms for each of the three experiments. All three algorithms begin similarly.  Both the held part and the reference part have a circular feature that can be used to locate them, so we deploy \textsc{CircleLocator} to get the 3D pose of the relevant circles.  At the end of each algorithm, we move the robot to a safe pose $\vec{x}_{safe}$.  This is done so that we can assume that every task begins with the robot in a known, safe configuration that far away from contacting either the base plate or staged parts.

Our bearing task algorithm begins by locating the bearing and the hole it belongs in.  Appropriate view locations from which the gripper can view the circular features of the bearing and hole must be provided the by user in $\vec{x}_{BrgView}$ and $\vec{x}_{HolView}$, respectively.  Then the bearing is grasped at the flange, with the gripper oriented vertically (fingers pointed $-Z$ down in the lab frame).  The bearing is then moved above the hole by the \textbf{InsertTilt} routine, and a tilted insertion is attempted.  Note that \textbf{InsertTilt} does not monitor the force-torque state of the wrist during the tilt-in operation, and the held part is tilted to the vertical regardless of what forces are sensed.  After the insertion attempt completes, then a tamping operation is initiated from above the center of the bearing hole location.

\begin{algorithm*}[!htb]
    \caption{Algorithm for the Bearing Insertion Task} \label{alg:bearing}
    \begin{algorithmic}[1]
        \Procedure{BearingTask}{ $\vec{x}_{BrgView}$ , $\vec{x}_{HolView}$ , Dia$_{Bearing}$ , Dia$_{Hole}$ , $z_{view}$ , $z_{PickDepth}$ , $\Delta z$ , $\theta_{tilt}$ , $F_{touch}$ , $F_{insert}$ , $\Delta_{max}$ , $d_{fingerSep}$ , $\vec{x}_{safe}$ }
        \State 
        
        \State \textbf{MoveAbs}( $\vec{x}_{BrgView}$ )
        \State $\vec{x}_{BrgLoc} \leftarrow$ \textsc{CircleLocator( Dia$_{Bearing}$ , $z_{view}$ )}
        
        \State \textbf{MoveAbs}( $\vec{x}_{HolView}$ )
        \State $\vec{x}_{HolLoc} \leftarrow$ \textsc{CircleLocator( Dia$_{Hole}$ , $z_{view}$ )}
        
        \State \textbf{Pickpart}( $\vec{x}_{BrgLoc}$ , Dia$_{Bearing}$ , $z_{PickDepth}$ )
        
        \State \textbf{InsertTilt}( $\vec{x}_{HolLoc}$ , $\Delta z$ , $\theta_{tilt}$ , Dia$_{Hole}$ , $\Delta x$ , $F_{touch}$ , $F_{insert}$ , $\Delta_{max}$ )
        
        \State \textbf{OpenGripper}()
        \State \textbf{SetGripperOpen}( $d_{fingerSep}$ )
        \State \textbf{TampWithForceLimit}( $\vec{x}_{HolLoc}$ , $\Delta z$ , $F_{touch}$ )
        
        \State \textbf{MoveAbs}( $\vec{x}_{safe}$ )
        
        \EndProcedure
    \end{algorithmic}
\end{algorithm*}

Likewise, our pulley task algorithm begins by locating the top of the shaft (reference part) and the pulley (moved part).  The pulley is grasped and brought to a position above the shaft with the plane of the pulley parallel with the plane of the base plate.  Then, a spiral insertion is attempted with the intent of sliding the pulley down onto the shaft.  Spiral insertion is followed by three tamping actions targeting the pulley; two on either side of the shaft, and one centered on the shaft.  Finally, a twist operation, centered on the shaft, is executed with the assumption that the pulley is jammed at the top of the shaft.  The twist operation does not monitor the gripper finger separation.  So, if twisting is initiated after the pulley has fallen to the base, then \textbf{Twist} will grasp and twist the shaft with no result.

\begin{algorithm*}[!htb]
    \caption{Algorithm for the Small Pulley Task} \label{alg:pulleyTask}
    \begin{algorithmic}[1]
        \Procedure{PulleyTask}{ $\vec{x}_{ShaftView}$ , $\vec{x}_{PullyView}$ , Dia$_{Shaft}$ , Dia$_{Pulley}$ , $z_{PickDepth}$  , $\Delta z$ , $F_t=1$ , $F_d=1$ , $F_i=2$ , $\Delta_{max}$ , $d_{TampOffset}$ , $F_{touch}$ , $d_{fingerSep}$ , $d_{offset}$ , $z_{grasp}$ , $\vec{x}_{safe}$ }
        \State 
        
        \State \textbf{MoveAbs}( $\vec{x}_{ShaftView}$ )
        \State $\vec{x}_{ShftLoc} \leftarrow$ \textsc{CircleLocator( Dia$_{Shaft}$ , $z_{view}$ )}
        
        \State \textbf{MoveAbs}( $\vec{x}_{PullyView}$ )
        \State $\vec{x}_{PulyLoc} \leftarrow$ \textsc{CircleLocator( Dia$_{Pulley}$ , $z_{view}$ )}
        
        \State \textbf{Pickpart}( $\vec{x}_{PulyLoc}$ , Dia$_{Pulley}$ , $z_{PickDepth}$ )
        
        \State \textbf{InsertPartSpiral}($\vec{x}_{ShftLoc}$, $\Delta z$, $F_t=1$, $F_d=1$, $F_i=2$, $\Delta_{max}$)
        
        \State \textbf{MoveAbs}( $\vec{x}_{safe}$ )
        \State \textbf{SetGripperOpen}( $d_{fingerSep}$ )
        \State \textbf{TampWithForceLimit}( $\vec{x}_{ShftLoc} + [0,-d_{offset},0,0,0,0]$ , $\Delta z$ , $F_{touch}$ )
        \State \textbf{TampWithForceLimit}( $\vec{x}_{ShftLoc} + [0, d_{offset},0,0,0,0]$ , $\Delta z$ , $F_{touch}$ )
        \State \textbf{TampWithForceLimit}( $\vec{x}_{ShftLoc}$ , $\Delta z$ , $F_{touch}$ )
        
        \State \textbf{Twist}( $\vec{x}_{ShftLoc}$ , $\theta_{osc}$ , $N_{osc}$ )
        
        \State \textbf{MoveAbs}( $\vec{x}_{safe}$ )
        
        \EndProcedure
    \end{algorithmic}
\end{algorithm*}

Our nut threading algorithm begins by locating the top edge of the stud (reference part) and the threaded hole of the nut (moved part).  The nut is grasped using the \textbf{PickPart} procedure described in Section \ref{placeProcs}.  The algorithm requires that two of the nut flats are parallel to the lab $Y$-axis in the lab reference frame in order to be gripped securely.  Also, the nut is intended to be placed on the stud in this same orientation such that the threading action grasps the nut by the flats. (Although, the threading action was shown capable of threading the nut to finger-tightness even when the nut was grasped by the corners.)  We attempt to seat the nut on top of the stud using a spiral insertion action.  The threads do not allow the nut to slide down the stud, but the taper on the stud and the nut do allow the nut to rest on the stud until threading can begin.

\begin{algorithm*}[!htb]
    \caption{Algorithm for the Nut Threading Task} \label{alg:nutTask}
    \begin{algorithmic}[1]
        \Procedure{NutThreading}{ $\vec{x}_{StudView}$ , $\vec{x}_{NutView}$ , Dia$_{Stud}$ , Dia$_{Hole}$ , Dia$_{Nut}$ , $z_{PickDepth}$  , $\Delta z$ , $F_t=4$ , $F_d=4$ , $F_i=2$ , $\Delta_{max}$ , $N_{flats}$ , $T_{z,limit}$ , $d_{pitch}$ $\vec{x}_{safe}$ }
        \State 
        
        \State \textbf{MoveAbs}( $\vec{x}_{StudView}$ )
        \State $\vec{x}_{StudLoc} \leftarrow$ \textsc{CircleLocator( Dia$_{Stud}$ , $z_{view}$ )}
        
        \State \textbf{MoveAbs}( $\vec{x}_{NutView}$ )
        \State $\vec{x}_{NutLoc} \leftarrow$ \textsc{CircleLocator( Dia$_{Hole}$ , $z_{view}$ )}
        
        \State \textbf{Pickpart}( $\vec{x}_{NutLoc}$ , Dia$_{Nut}$ , $z_{PickDepth}$ )
        
        \State \textbf{InsertPartSpiral}($\vec{x}_{StudLoc}$, $\Delta z$, $F_t=4$, $F_d=4$, $F_i=2$, $\Delta_{max}$)
        
        \State \textbf{MoveAbs}( $\vec{x}_{safe}$ )
        \State \textbf{OpenGripper}()
        
        \State \textbf{Thread}( $\vec{x}_{StudLoc}$ , $N_{flats}$ , $T_{z,limit}$ , $d_{pitch}$ )
        \State \textbf{OpenGripper}()
        \State \textbf{MoveAbs}( $\vec{x}_{safe}$ )
        
        \EndProcedure
    \end{algorithmic}
\end{algorithm*}

\section{Experiment} 

The three tasks described in Section \ref{taskDesc} were repeated 20 times each.  For each of the trials, wrist forces, wrist torques, effector pose, and joint configuration were recorded.  Each trial was manually labelled as either a success or a failure based upon whether the moved part reached its intended configuration.  The bearing task is a success if it is seated such that the flange is resting on the base plate.  The pulley task is a success if the pulley is on the shaft, the pulley is resting on the base plate, and the pulley rotates freely about the shaft.  The nut threading task is a success if the nut is threaded onto the stud, and the nut is touching the base plate.  Tightness of the nut is not a determinant of success.  Rotation about the lab reference $Z$-axis was not considered when determining if the moved parts were properly assembled.

In order to test the effectiveness of the hole feature detection capability, we selected three holes on the task board; $9$mm, $17$mm, and $35$mm in diameter. We then move the camera through $25$ points regularly spaced on a square grid $100$mm per side. At each point we use the system to locate the target circle, and record success or failure. This was done at $70$mm, $170$mm, and $270$mm. This entire process was repeated 3 times, and the results are shown in Table \ref{successHole}.

\section{Results} \label{sct:results} 

The success rate of each task is shown in Table \ref{successTable}.  The mean times to completion or failure for each task are reported in Table \ref{TimesTable}. 

\begin{figure}[t]
    \centering
    \captionof{table}{Task Success Rates}	\label{successTable}
    \begin{tabular}{|l|r|r|r|} \hline
        
        \textbf{Task} & \textbf{Trials} & \textbf{Success} & \textbf{Rate} \\ \hhline{|=|=|=|=|}
		
		Large Bearing & 20 & 13 & 0.65 \\ \hline
		
		Small Pulley  & 20 & 13 & 0.65 \\ \hline
		
		Stud          & 20 &  7 & 0.35 \\ \hline

    \end{tabular} 
\end{figure}

\begin{figure}[t]
    \centering
    \captionof{table}{Mean Running Times [s]}	\label{TimesTable}
    \begin{tabular}{|l|r|r|} \hline
        
        \textbf{Task} & \textbf{Success} & \textbf{Failure} \\ \hhline{|=|=|=|}
		
		Large Bearing &  20.0 & 18.9 \\ \hline
		
		Small Pulley  &  79.9 & 48.0 \\ \hline
		
		Stud          & 175.3 & 41.1 \\ \hline

    \end{tabular} 
\end{figure}

\begin{figure}[b]
    \centering
    \captionof{table}{{Hole Detection Success Rates}}	\label{successHole}
    \begin{tabular}{|l|r|r|r|} \hline
        
        \textbf{Hole Dia} & \textbf{$7$cm height} & \textbf{$17$cm height} & \textbf{$27$cm height} \\ \hhline{|=|=|=|=|}
		
		$9$mm  & $0.92$ & $0.99$ & $1.00$ \\ \hline
		
		$17$mm & $0.59$ & $1.00$ & $1.00$ \\ \hline
		
		$35$mm & $0.96$ & $1.00$ & $1.00$ \\ \hline

    \end{tabular} 
\end{figure}

\subsection{Bearing Task} 

\begin{figure*}[!htb]
	\centering
	\includegraphics[height=1.3in]{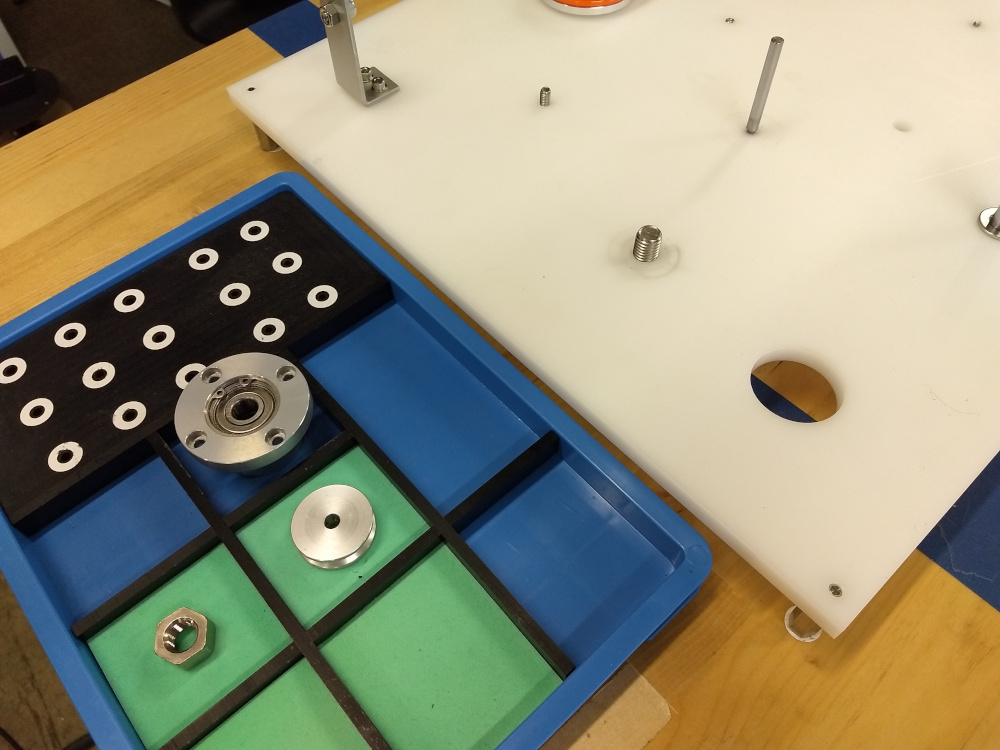}
	\includegraphics[height=1.3in]{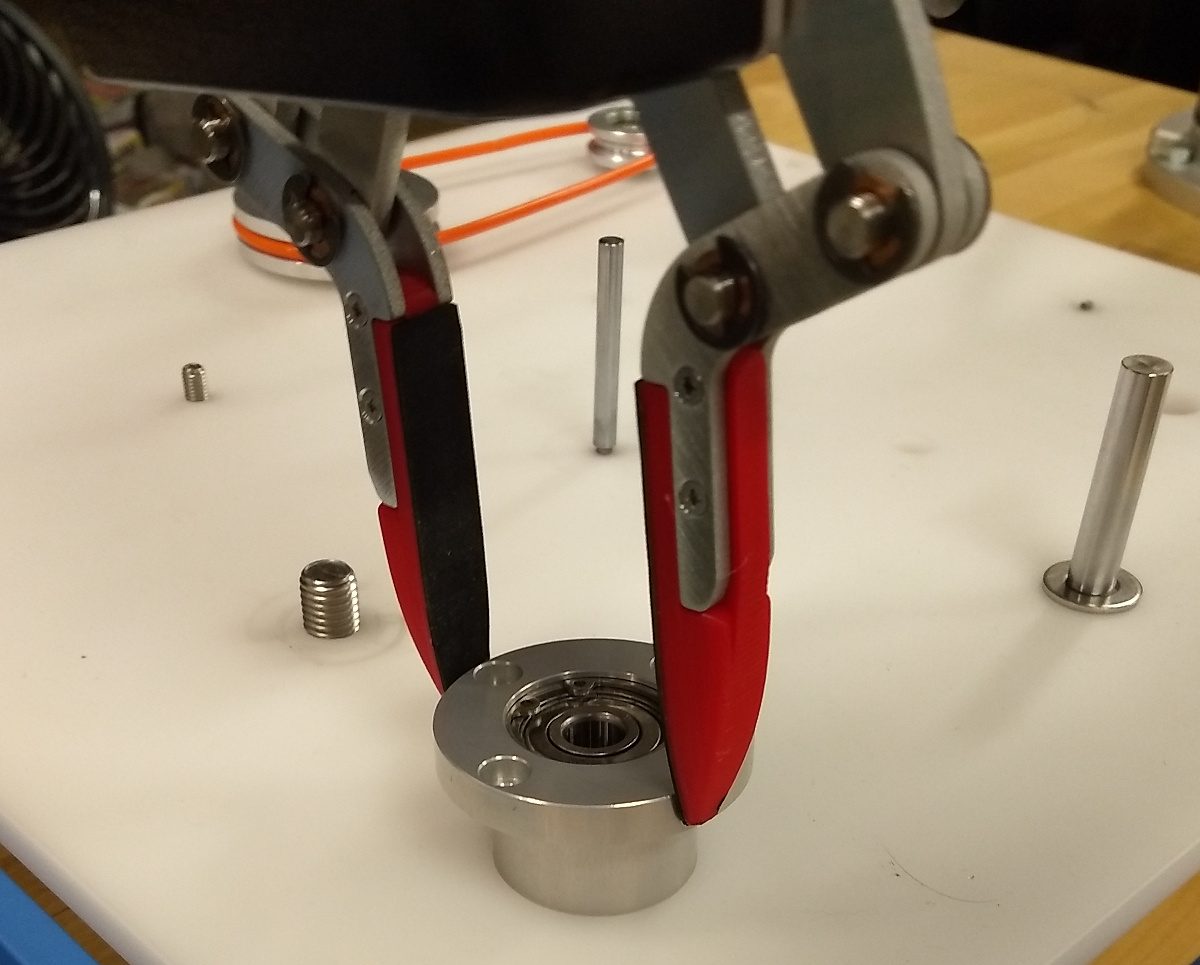}
	\includegraphics[height=1.3in]{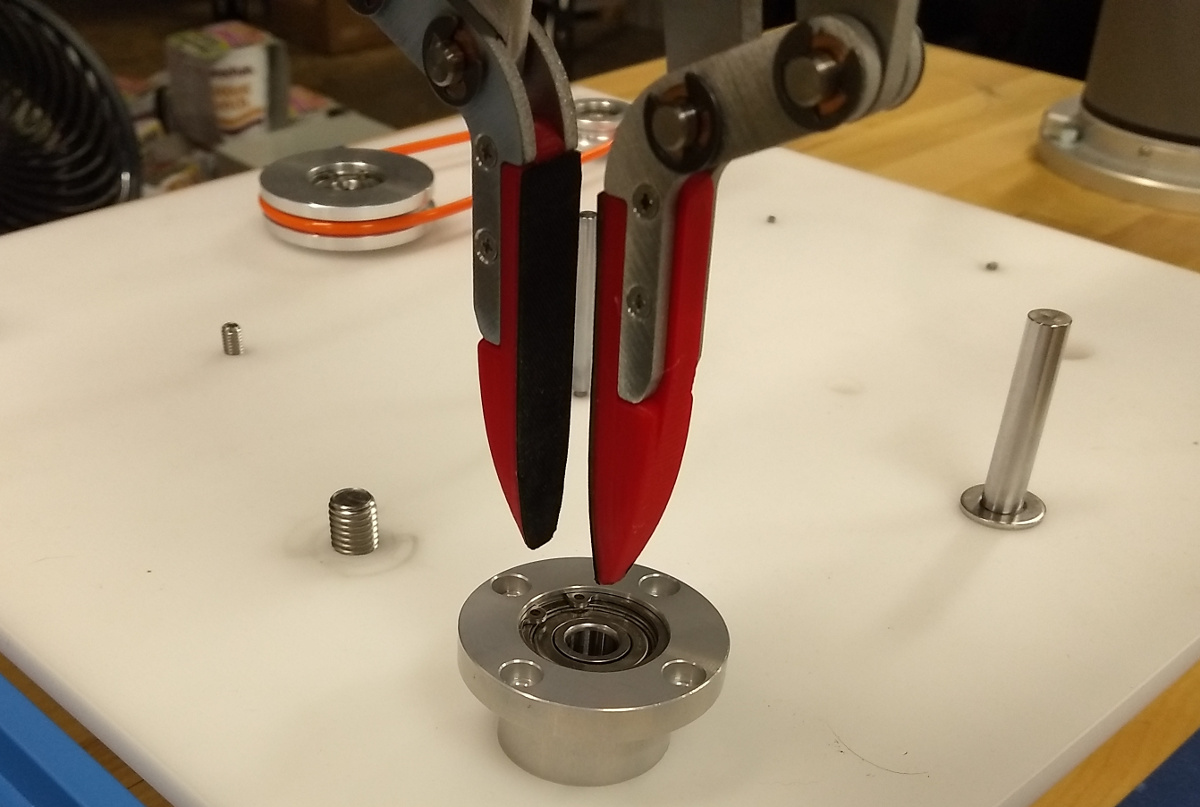}
	\caption{From left to right: experimental setup just prior to assembly with the bearing in the kitting tray. After picking the bearing, tilting insertion is used to insert the bearing. Finally, the robot presses down to complete insertion. }
	\label{photoBefore}\label{bearingForceSucceed}
\end{figure*}



The larger bearing has a relatively tight tolerance ($+0.2mm$) with its hole in the base plate, and its primary mode of failure is jamming.  A jam occurred while inserting the bearing on seven of the twenty attempts.  The tamp operation was able to seat the bearing in one of these cases, resulting in successful insertions.  In one of the seven failures, an unknown movement error occurred that caused the robot arm to miss the hole in the base plate.

\begin{figure*}[!htb]
	\includegraphics[height=1.6in]{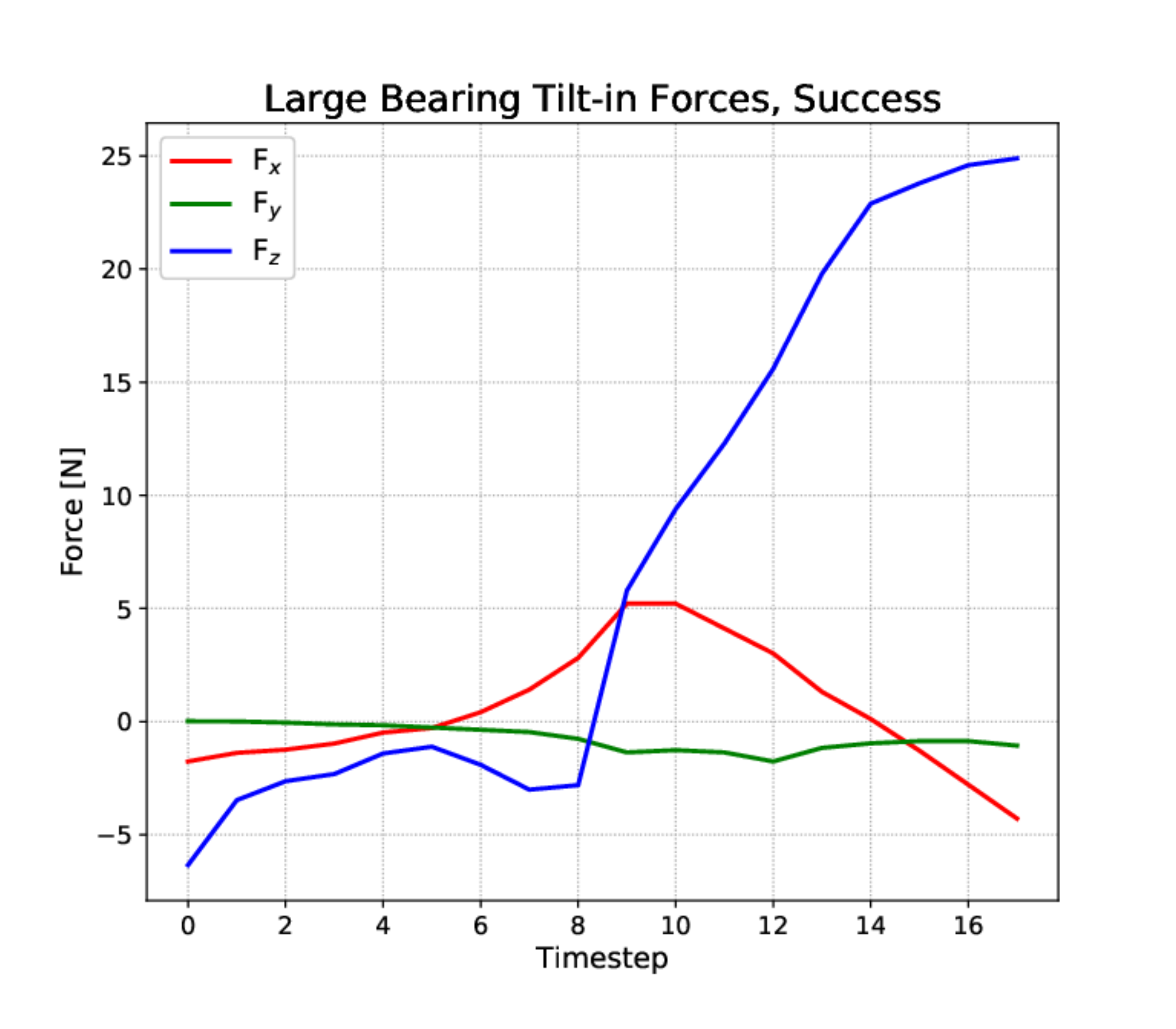}
	\includegraphics[height=1.6in]{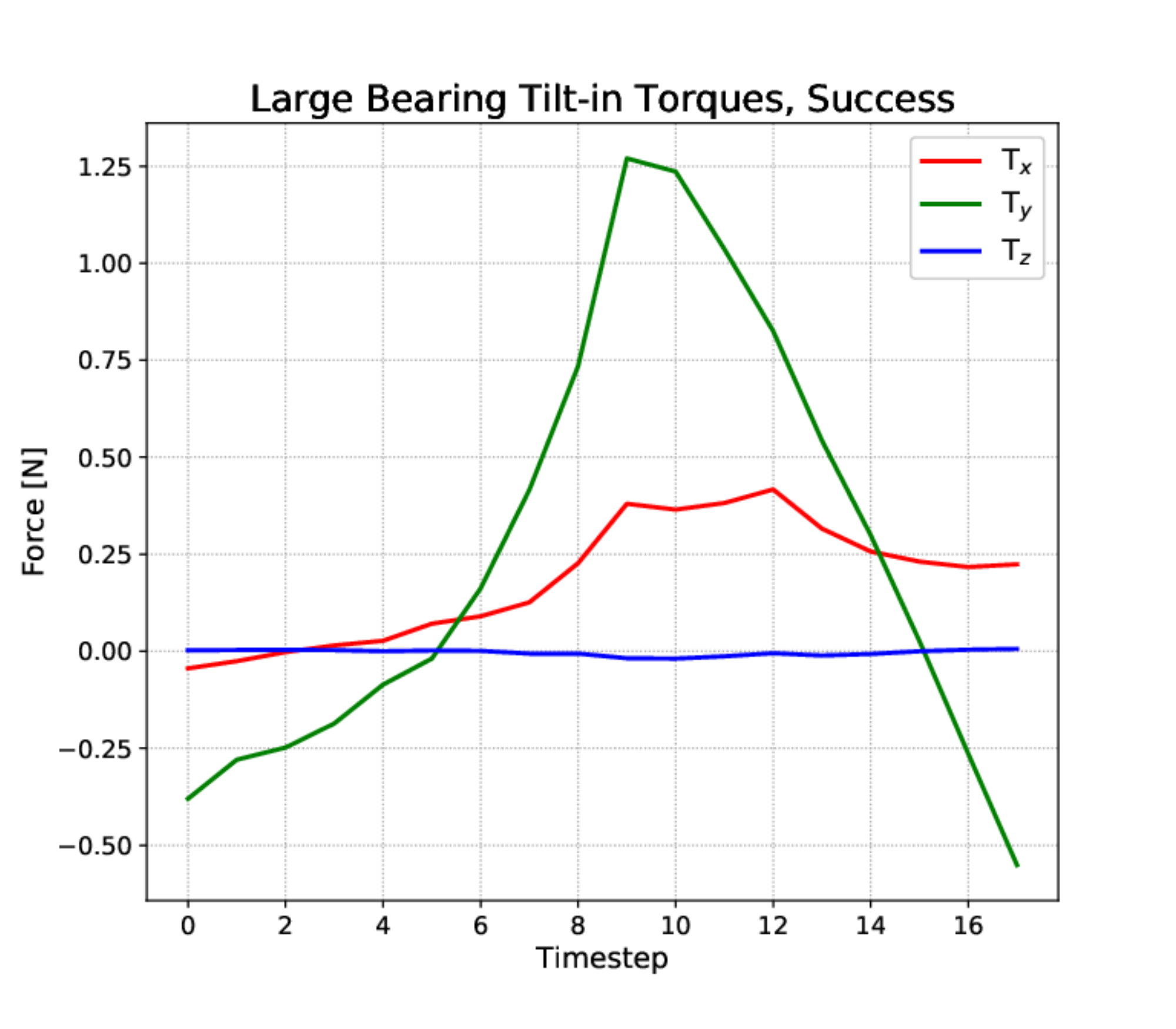}
	\includegraphics[height=1.6in]{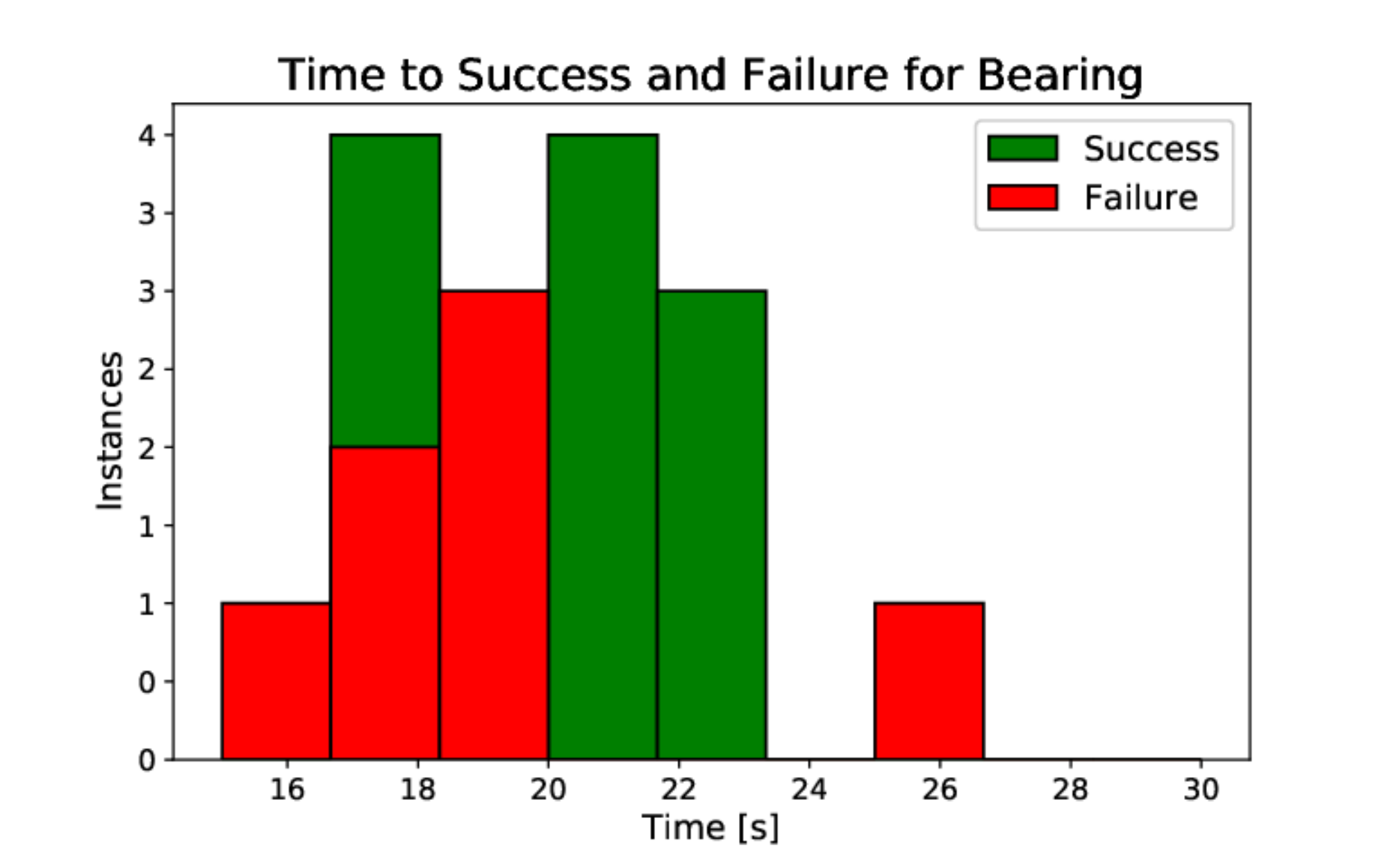}
	\\
	\includegraphics[height=1.6in]{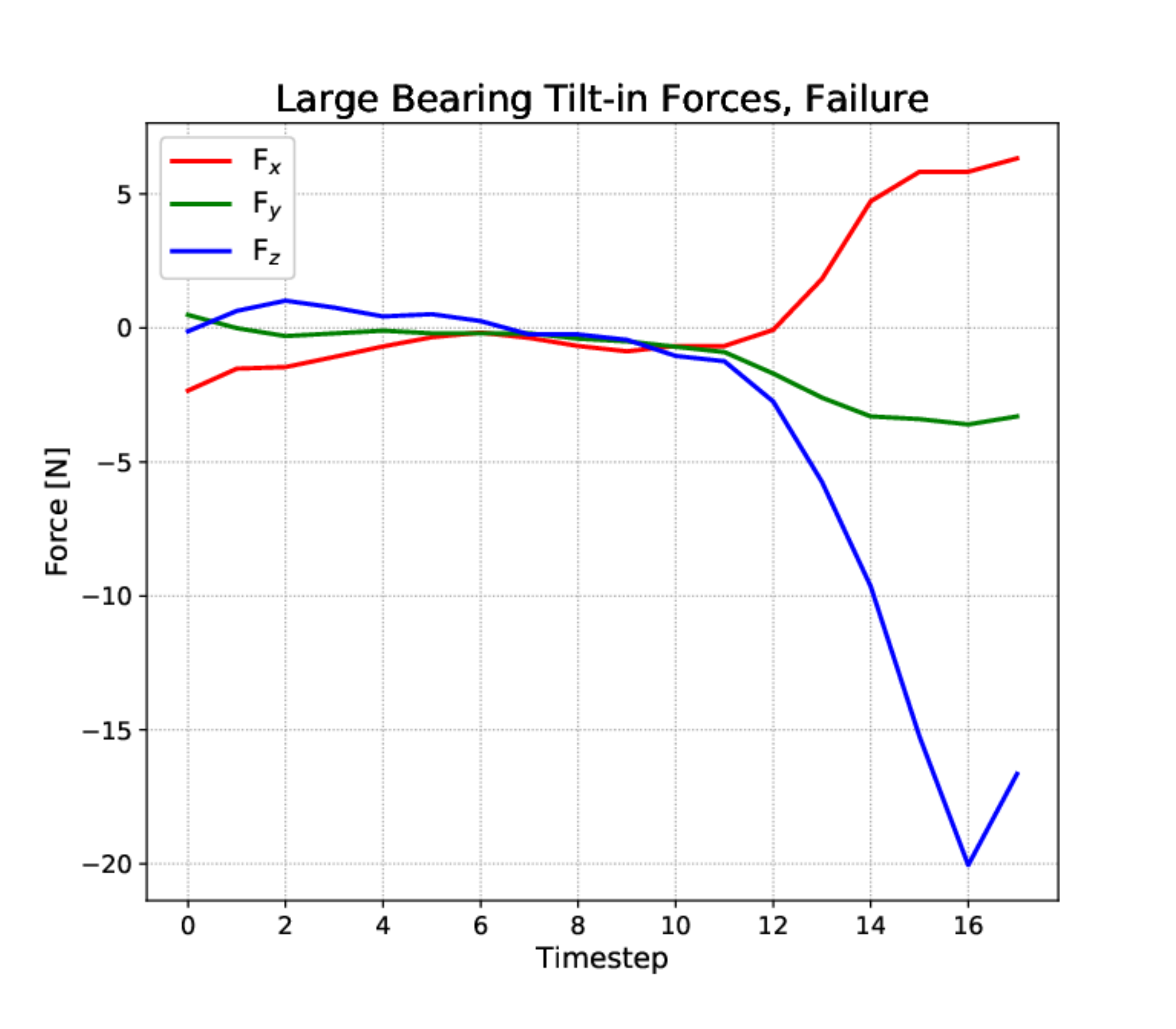}
	\includegraphics[height=1.6in]{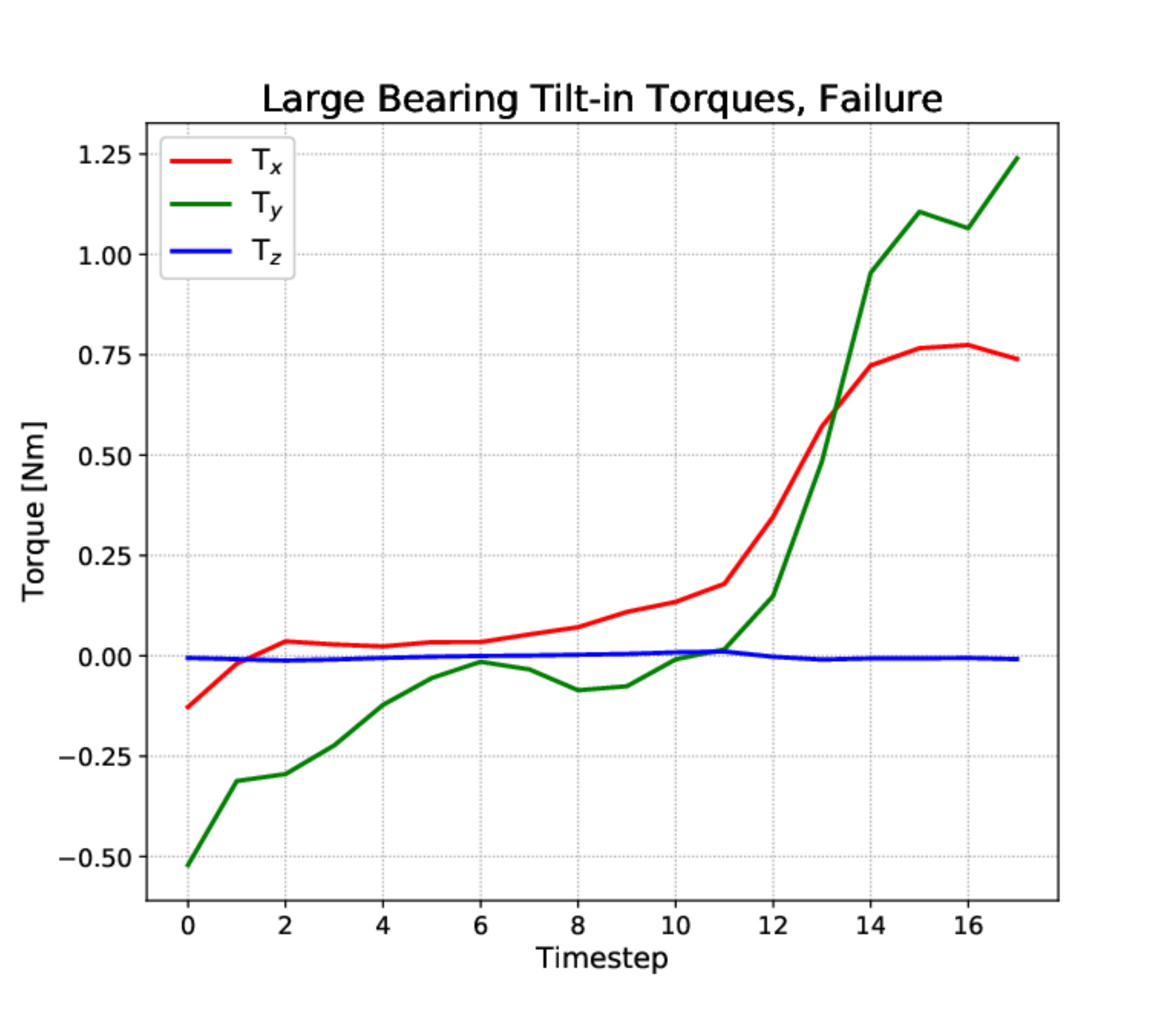}
	\caption{Sample wrist forces (left) and torques (center) vs. time during successful (top) and failed (bottom) bearing insertion. Time distribution for successes (green) and failures (red) are shown in the top right, 20 trials total.}
	\label{bearingForceSucceed}
	\label{bearingTorqueSucceed}
	\label{bearingForceFailure}
	\label{bearingTorqueFailure}
	\label{bearingHisto}
\end{figure*}





The force profile (Figure \ref{bearingForceSucceed}, top left) for a successful bearing insertion shows a large positive increase in $F_z$ (away from the wrist). This is due first to the bearing falling into the hole in the base, then to the bearing being bound in the hole as the robot wrist completes its arc to the vertical orientation.

The opposite trend in $F_z$ can be seen in force profile of an unsuccessful trial (Figure \ref{bearingForceFailure}, bottom left).  The robot tries to tilt the bearing into the hole, but the bearing is not aligned properly, and the bearing runs into the base plate at the edge of the hole, resulting in the reaction force in $-F_z$ (towards the wrist.) 

During the bearing insertion operation, the robot hand is rotating about its local $Y$-axis, and the angle between the axis of the bearing and the axis of the hole in the base decreases.  As the bearing tilts up, its contact point with the edge of the hole get further from the center of rotation, so there is greater resistance torque to further rotation until the bearing clears the edge of the hole at the top of the $T_y$ curve as shows in Figure \ref{bearingTorqueSucceed}, top right.  In the failure case, $T_y$ continues to increase and bearing never enters the hole in the base plate. (Figure \ref{bearingTorqueFailure}, bottom right).

The mean times to failure and success for the bearing task are similar, as the tilt-in action takes up the majority of the running time (Figure \ref{bearingHisto}).  All observed failures occurred during the insertion phase.  Failure is immediately apparent by the end of the insertion action because the bottom edge of the bearing has either cleared the edge of its base plate hole, or it has not.  In the former case, the tamping action that follows insertion cannot correct the condition.  The mean time to success is slightly longer than that to failure because some successes are the result of the tamping action.  There is one outlier failure case in which the bearing edge clears the hole, but the bearing is so badly jammed that it cannot be dislodged by tamping.

Van Wyk \textit{et al} \cite{van2018comparative} use a tilt-insert method very similar to ours with a four-digit robot hand and in-finger load cells, albeit using much larger 3D-printed parts and hard-coded positions instead of vision. The system they present achieves a slightly higher success rate ($0.88$) than ours, and signifies there is a great opportunity to improve on the benchmark presented here.

\subsection{Pulley Task} \label{pulleyFails} 

\begin{figure*}[!htb]
	\centering
	\includegraphics[height=1.3in]{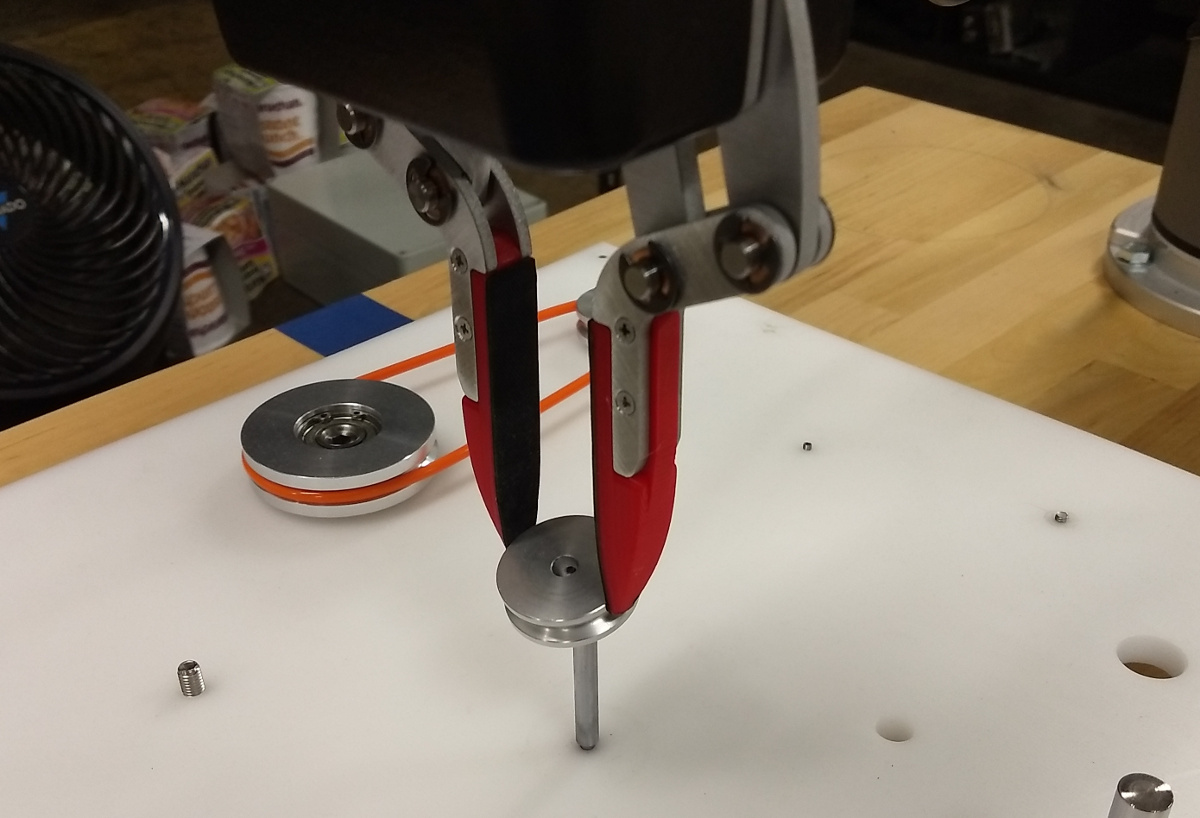}
	\includegraphics[height=1.3in]{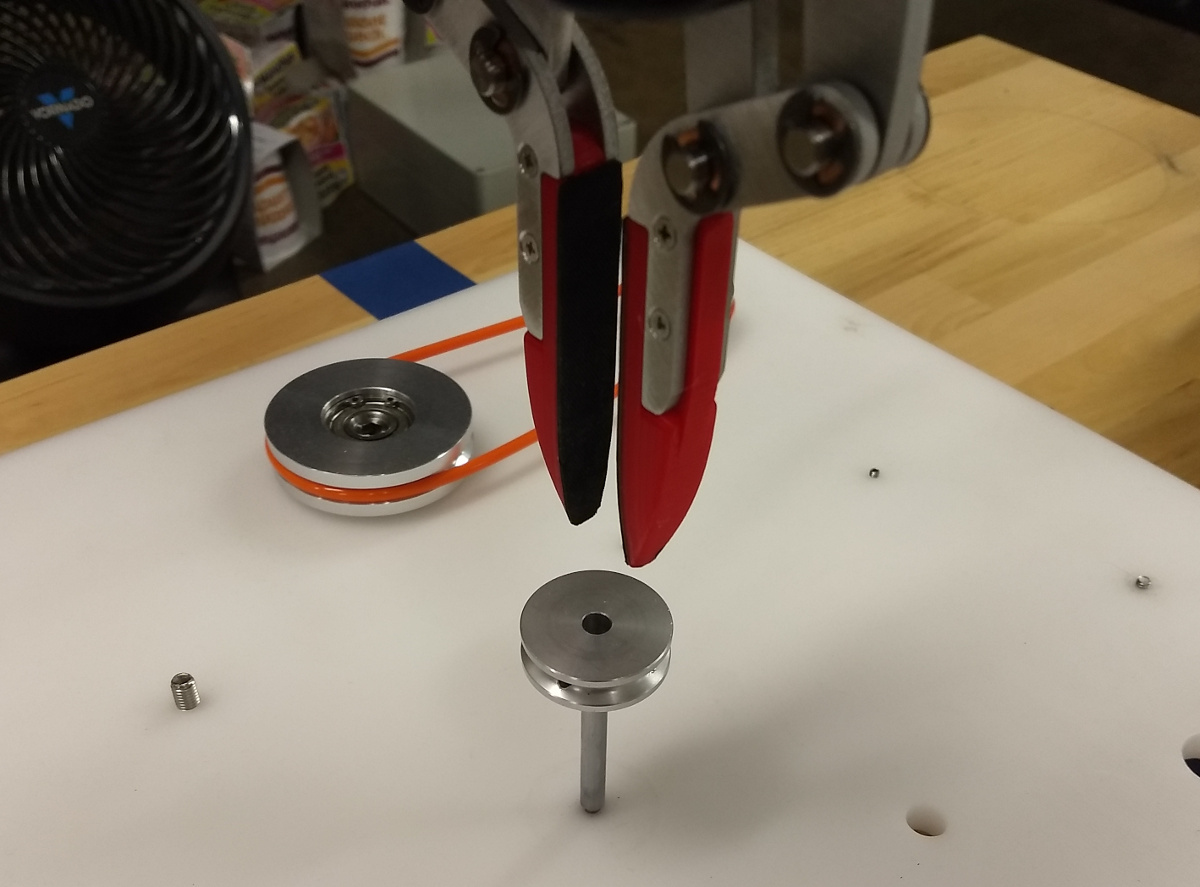}
	\includegraphics[height=1.3in]{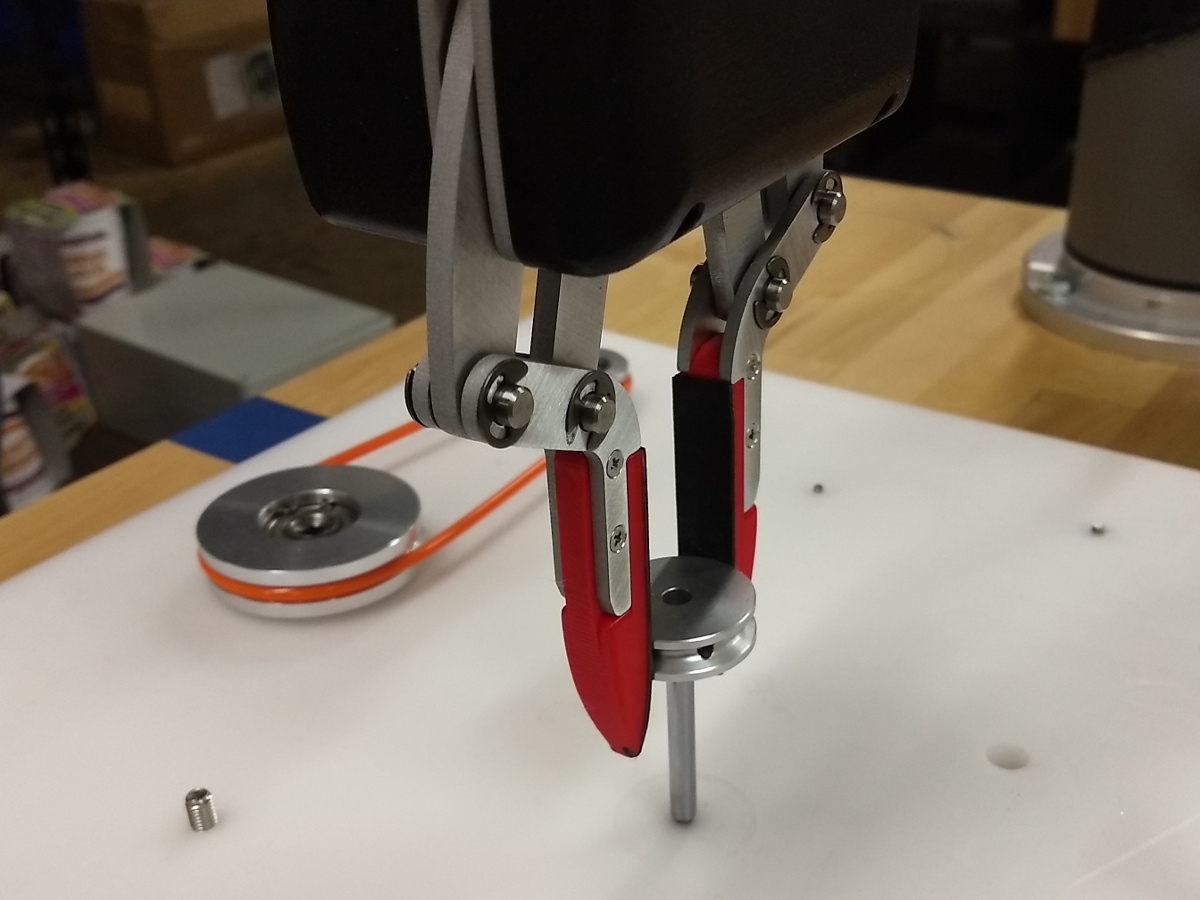}
	\caption{Steps during pulley assembly, from left to right: approach using a spiraling motion, pushing down, final twist motions.}
	\label{fig:pulleyassembly}
\end{figure*}




\begin{figure*}[!htb]
	\includegraphics[height=1.6in]{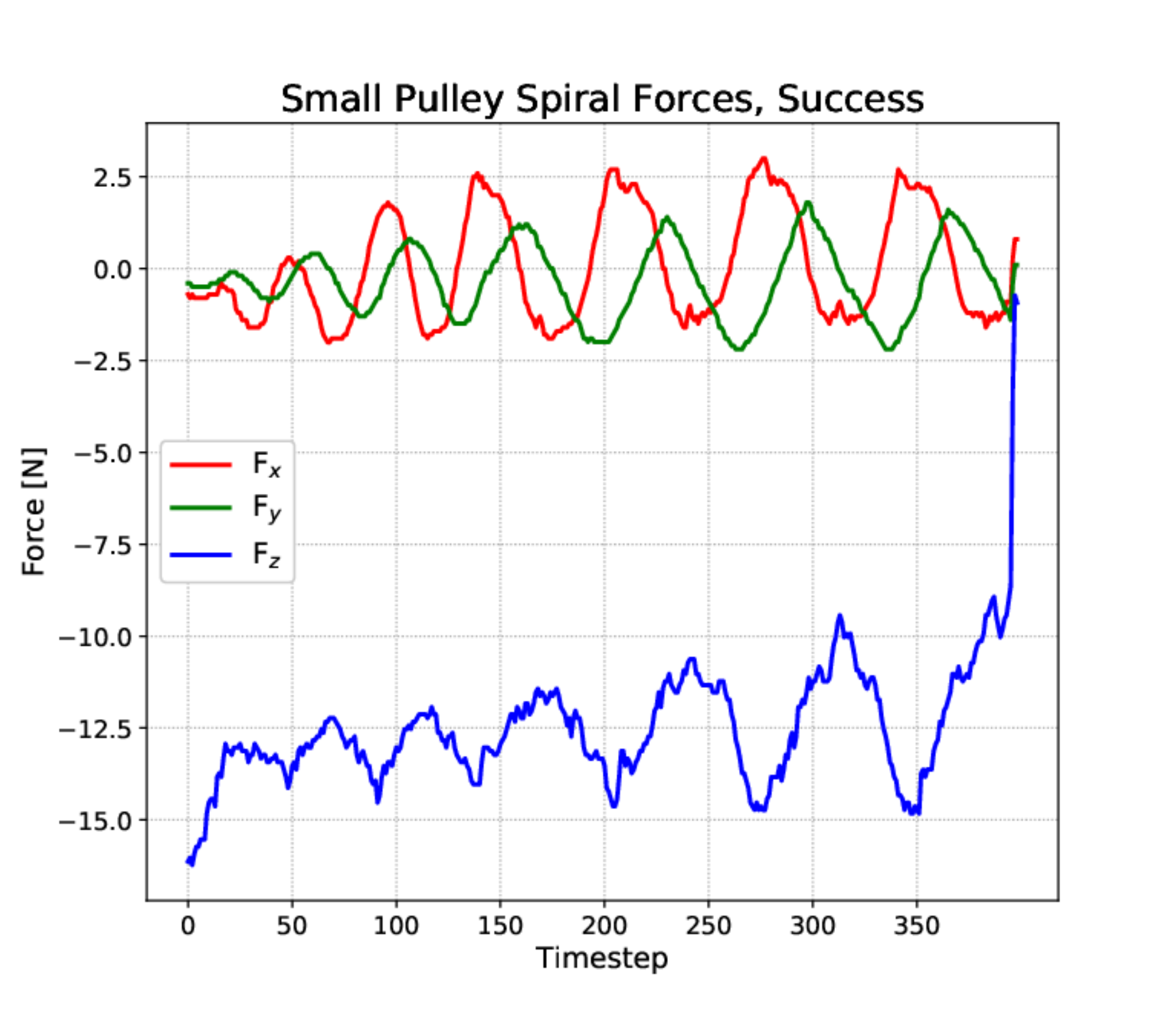}
	\includegraphics[height=1.6in]{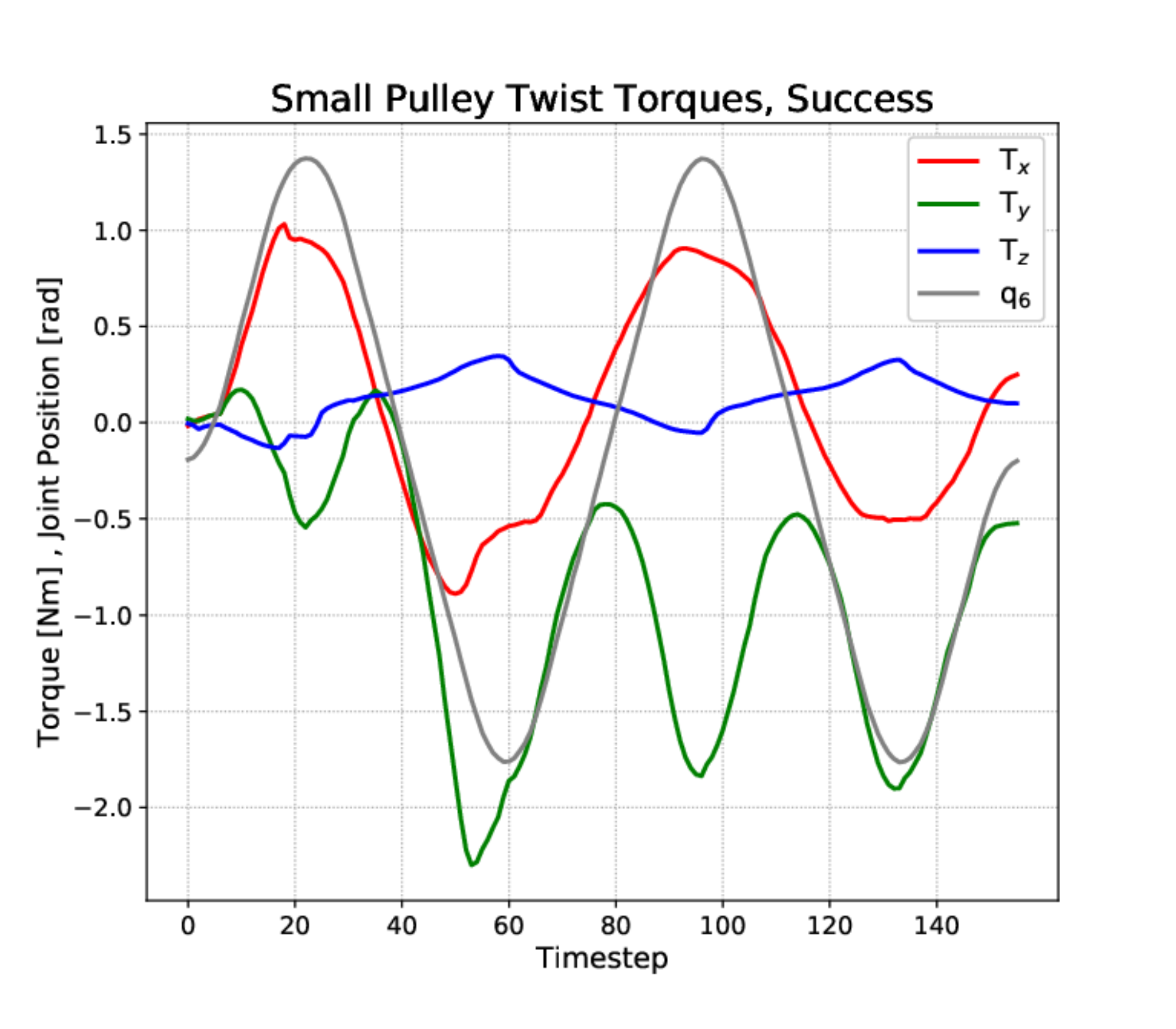}
	\includegraphics[height=1.6in]{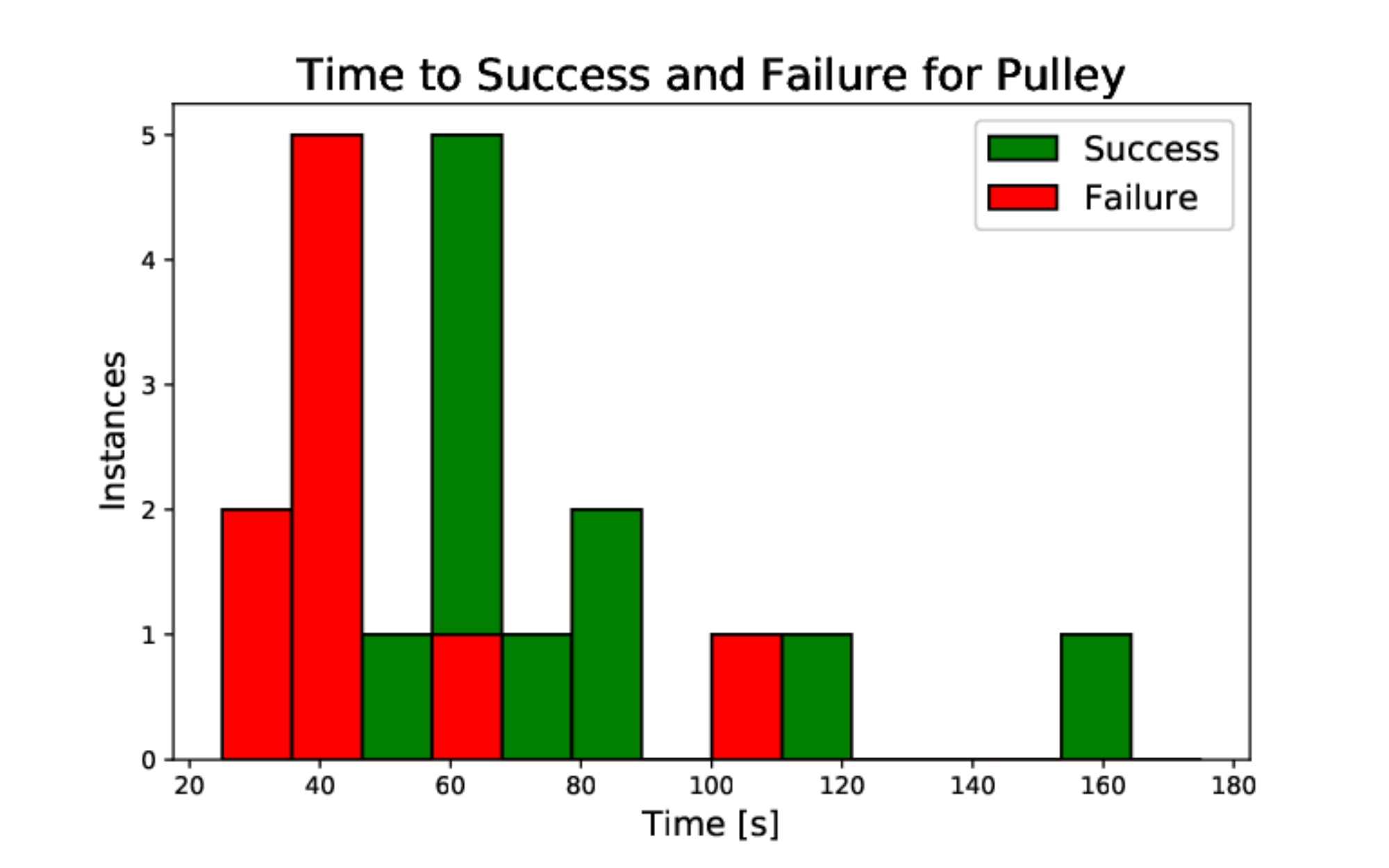}\\
	\includegraphics[height=1.6in]{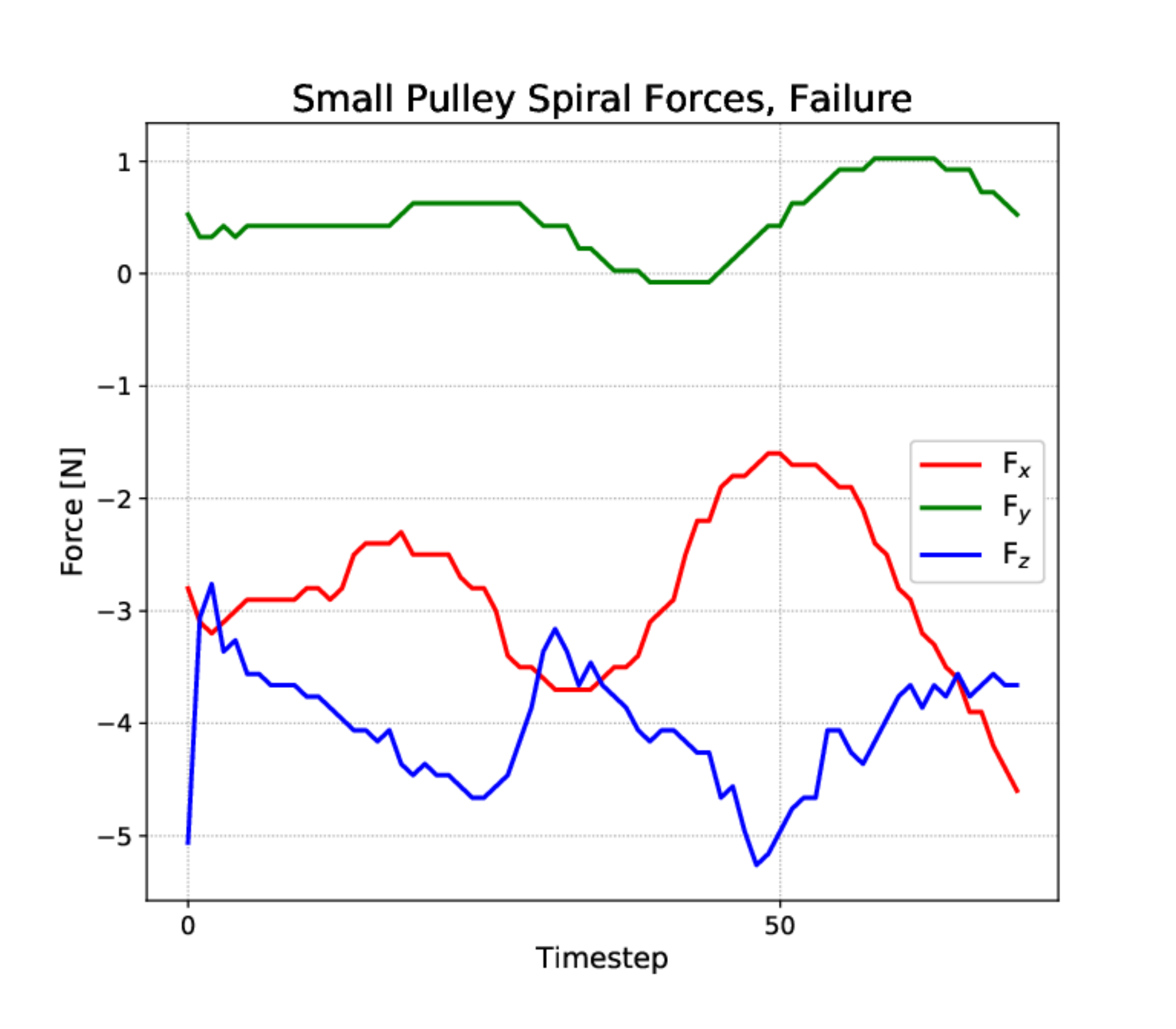}
	\includegraphics[height=1.6in]{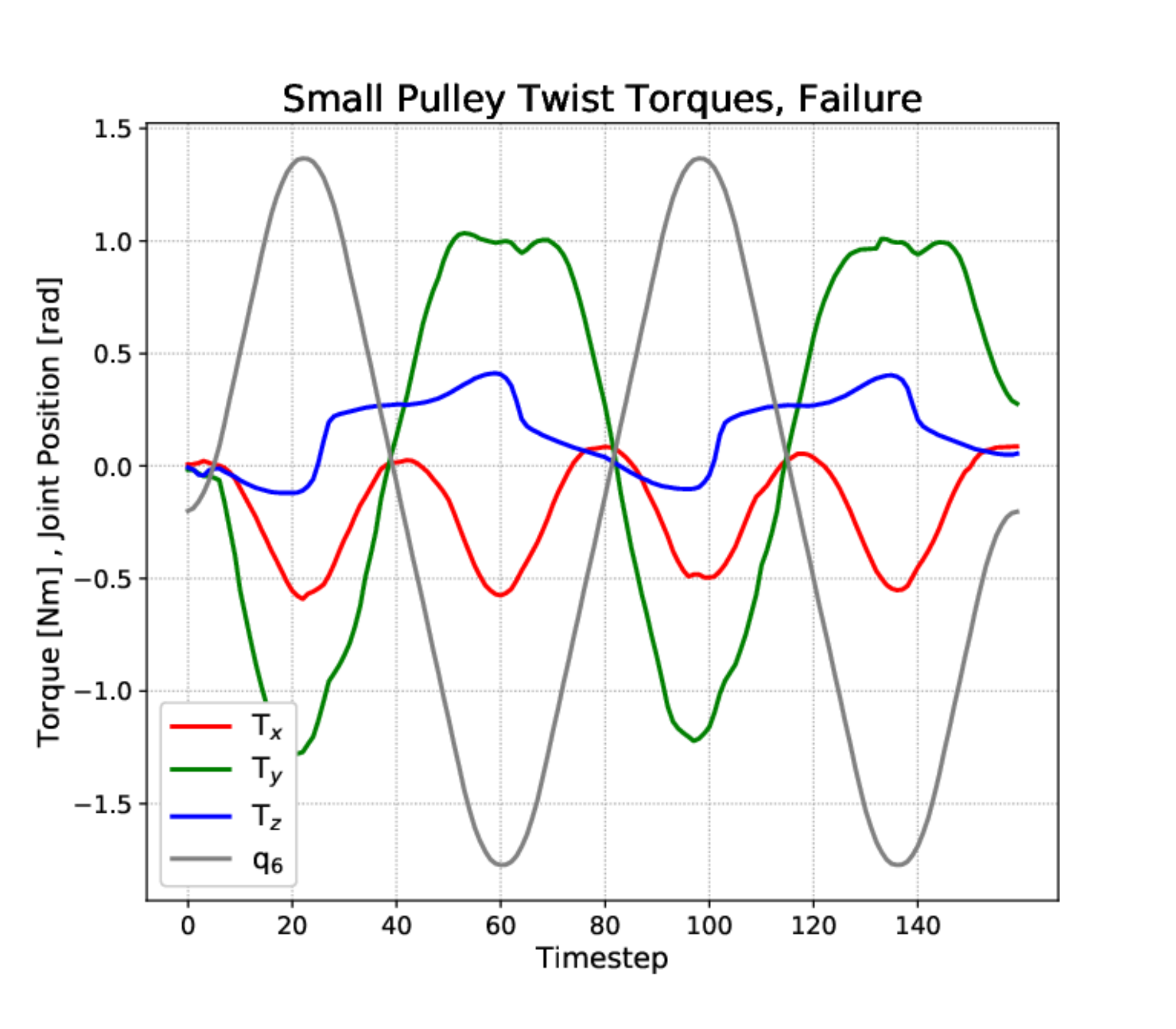}
\caption{Sample wrist forces (left) and torques (center) vs. time during successful (top) and failed (bottom) pulley assembly. Time distribution for success (green) and failures (red) are shown in the top right, 20 trials total. }
\label{pulleyForceSucceed}
\label{pulleyTorqSucceed}
\label{pulleyForceFail}
\label{pulleyTorqFailure}
\end{figure*}

A successful spiral insertion operation (Figure \ref{fig:pulleyassembly}) for the small pulley is shown in Figure \ref{pulleyForceSucceed}, top left.  The $X$ and $Y$ forces are similar in magnitude and 90 degrees out of phase, as expected.  The $Z$ force drops off suddenly when the pulley shaft clears the hole.

Failures of the spiral insertion operations witnessed during the experiment were typically due to the pulley being brought down onto the shaft in a way that caused the pulley to tilt in the fingers.  We can see the evolution of the operation in this case in Figure \ref{pulleyForceFail}, bottom left.  There is no resistance to motion in the $-Y$ direction, and very little in $+Y$. As the pulley moves in the spiral, the shaft touches the hole at $X$ extremities only and the stop condition is met without any opportunity to bring the pulley back into a horizontal orientation that allows insertion.  When the pulley is tilted this way, it will fall from the shaft as soon as the robot releases it.  This was observed in all 7 failed trials.

Even after a successful spiral operation, the most frequent outcome is for the pulley to be jammed on the shaft, even though the top edge of the shaft has cleared the hole.  The pulley ended the spiral operation in a jammed state in 12 out of 13 successful trials.  Several tamping operations are performed to attempt to recover from a jammed state.  The wrist sensor either experiences a $-Z$ spike in force when it contacts the jammed pulley, or the pulley has been freed and there is no contact.  Unfortunately the release of the jammed state does not have a recognizable profile and we must rely on the following tamp operation to detect the unjammed state.

The wrist torques observed during the twist action follow a oscillatory profiles with frequencies that are either the same as, or twice that of, the frequency of wrist motion.  The relationship between torques and wrist motion can be seen by comparing the top and bottom rows (center column) in Figure \ref{pulleyTorqSucceed}.  A failed twist action (bottom row) has no change in waveform throughout the action.  However, there is a noticeable step offset that manifests in the $Y$-torque felt at the wrist when a twist action corrects a jam.  This is due to axis of the pulley coming into alignment with the shaft.  When the twist action begins, the hand has grasped the pulley in its jammed, tilted configuration.  When the pulley aligns with he shaft, it rotates about the hand's $Y$-axis and imparts a constant torque to the hand about that axis.

Performance on the Small Pulley task had a lower success rate than the example set by Van Wyk \textit{et al} \cite{van2018comparative}, who were able to achieve a higher success ($0.95$) rate on a similar peg-in-hole task using a spiral search strategy, albeit using much larger 3D-printed plastic parts and hard-coded positions instead of vision.  



The mean time to failure of the pulley task was about 30 seconds less than the average time of completion.  Typically, the failure was in the spiral task, and the pulley would fall from the shaft, preventing any further actions from being taken, see also the histogram in Figure \ref{pulleyHisto}, right.  Task completion takes more time because in almost every success case, the spiral action ends with the pulley securely on the end of the shaft, but bound there due to some misalignment between the pulley and the shaft.  In most cases two to three tamps are required to dislodge the pulley and complete the task. There are two outliers, which represent the two trials that required a twist action to dislodge the pulley.

The initial approach to this task was unsuccessful, with all attempts resulting in failure.  There were two failure modes.  The first is misalignment, in which the spiral insertion action stopped when the lateral force criterion was met, but the pulley was tilted about an an axis formed by the gripper contact points.  This failure mode revealed a case in which the assumption of the stopping criteria was incorrect.  The second failure mode was jamming.  In this mode, the pulley shaft clears the edge of the pulley hole, and so the stopping force criterion of the spiral motion is met.  However, the hole and pulley axes are not aligned, and the pulley cannot slide down to the base plate.  At present we do not have a technique for compensating for the first failure mode, as our system does not have means of sensing tilt of the held part within the fingers.  Further investigation is needed for this case.  There were two compensatory actions developed in order to compensate the second failure mode.  The first is a twisting action that rotates the pulley about the shaft axis in an oscillatory way.  This was meant to mimic the way a human might free a jammed peg-in-hole operation, but was not successful in any of the robot trials attempted. The second compensatory action is a ``tamp'' that applies force on the moved part (pulley) up to a predefined limit.  Naturally, this has the potential to worsen a jammed condition if it imposes a moment on the pulley in the direction of its rotation out of the vertical.  In order to account for this, three successive tamps were applied in the along a horizontal axis perpendicular to the direction of the finger opening: two offset from the shaft center on opposite sides, and one over the shaft center.  This action was unsuccessful in resolving jams.  Several different pulleys were tried with the same shaft.  Neither the shaft nor the pulleys showed any obvious signs of defect or surface roughness on their contact surfaces.

The close tolerance between the pulley and the shaft made the jamming problem too persistent for the present system to compensate for.  This issue was relieved by applying a small amount of silicone lubricant to the contact surfaces.  This is not unreasonable, as it is expected for a product to ship with lubricant already applied to its moving parts.  However, it does highlight the need for the system to be able to detect such jammed conditions.  After lubrication, the jammed condition only occurred once during trials.  All other failures observed were of the misalignment type described in at the beginning of Section \ref{pulleyFails}. 

\subsection{Stud Task} \label{studEvolv} 

\begin{figure*}[!htb]
	\centering
	\includegraphics[height=1.3in]{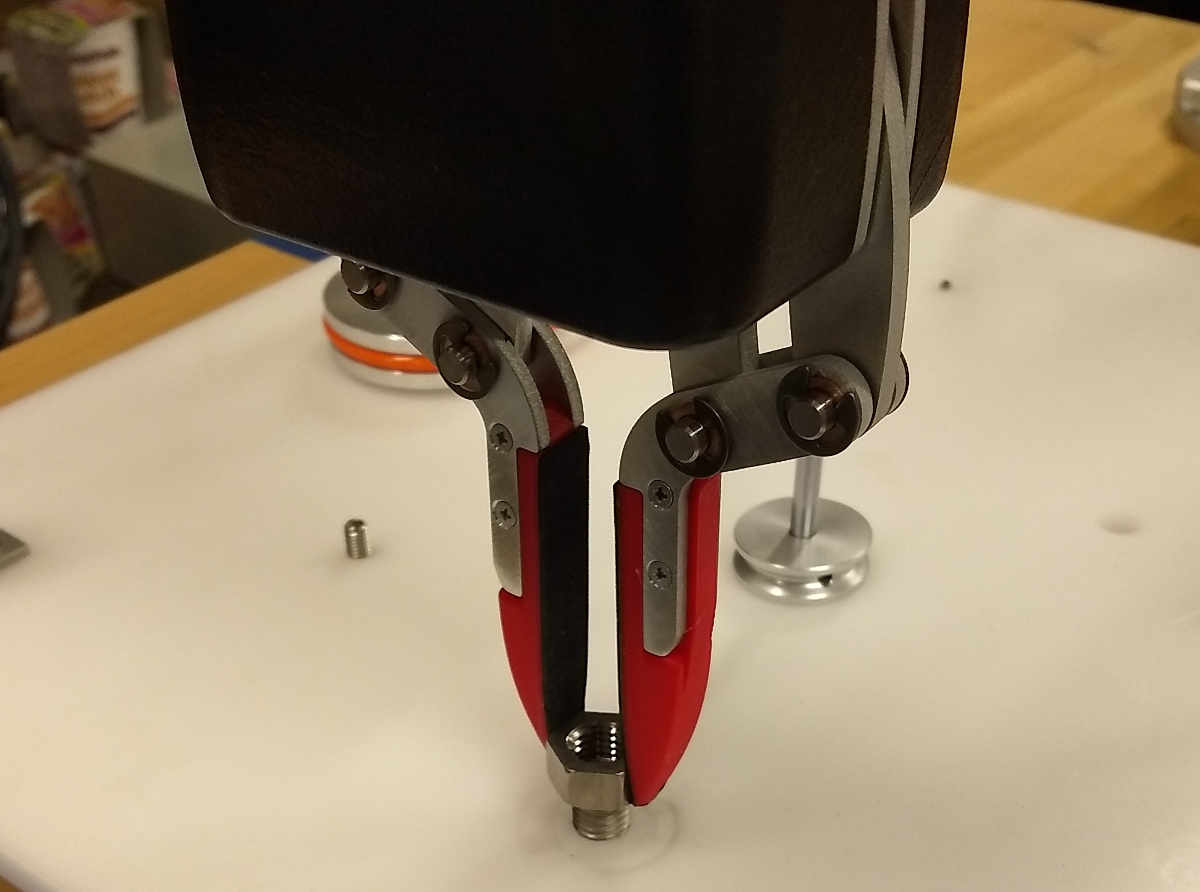}
	\includegraphics[height=1.3in]{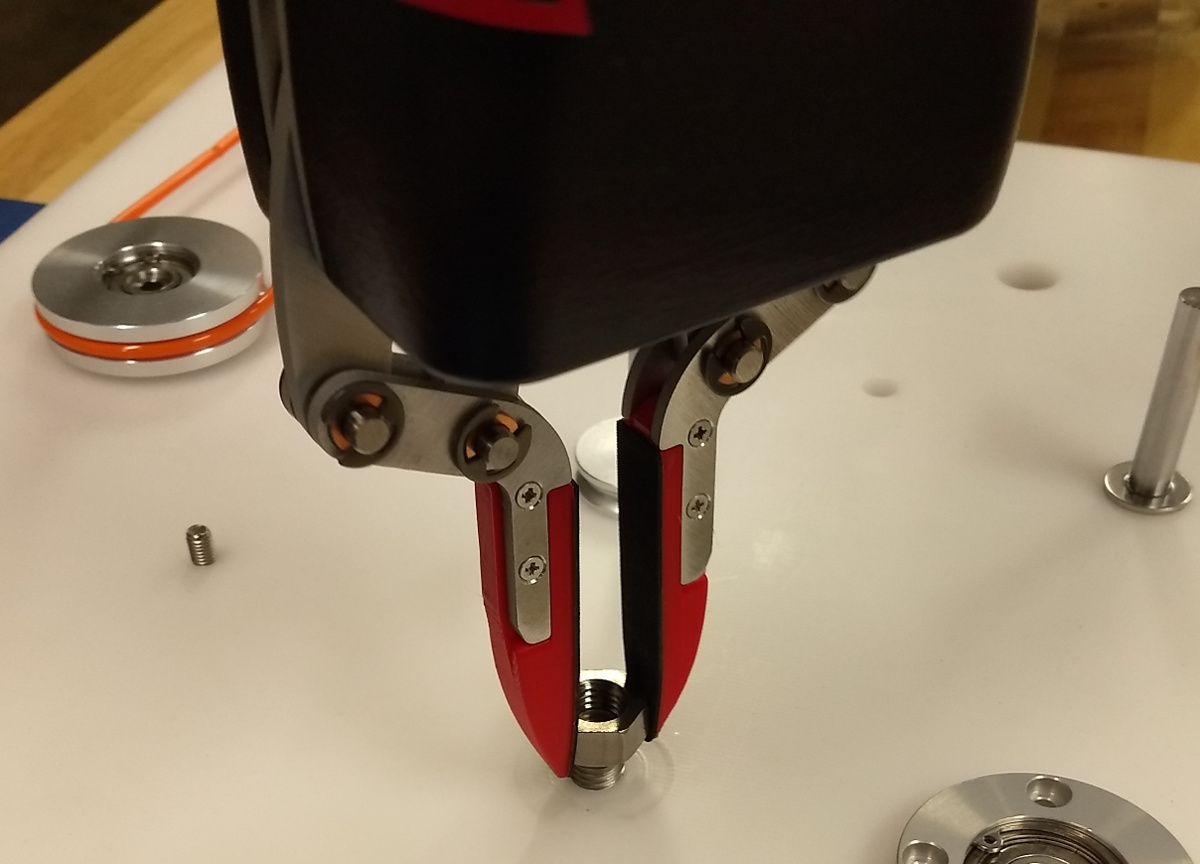}
	\includegraphics[height=1.3in]{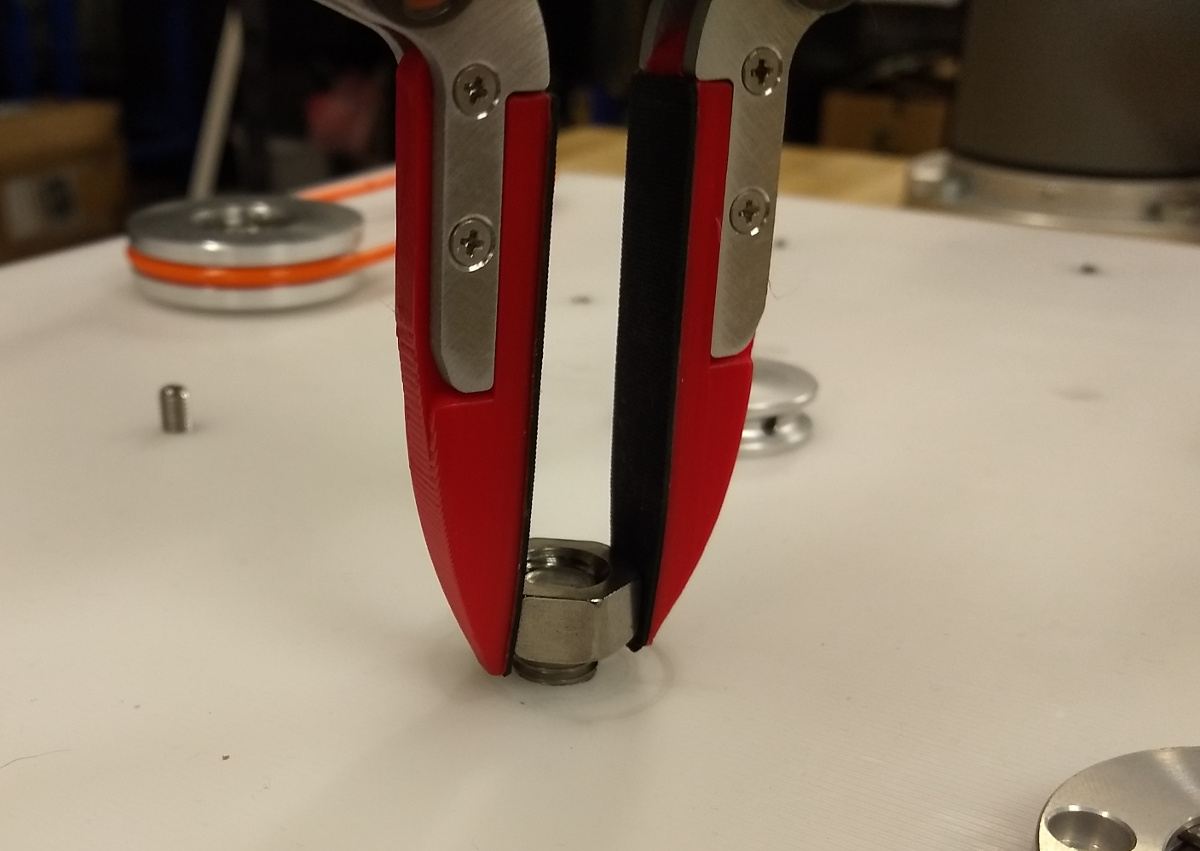}
	\caption{Steps during nut assembly, from left to right: approach using a spiraling motion, initial thread, final twist motion together with downward motion.}
	
\end{figure*}

\begin{figure*}[!htb]
\centering
\includegraphics[height=1.6in]{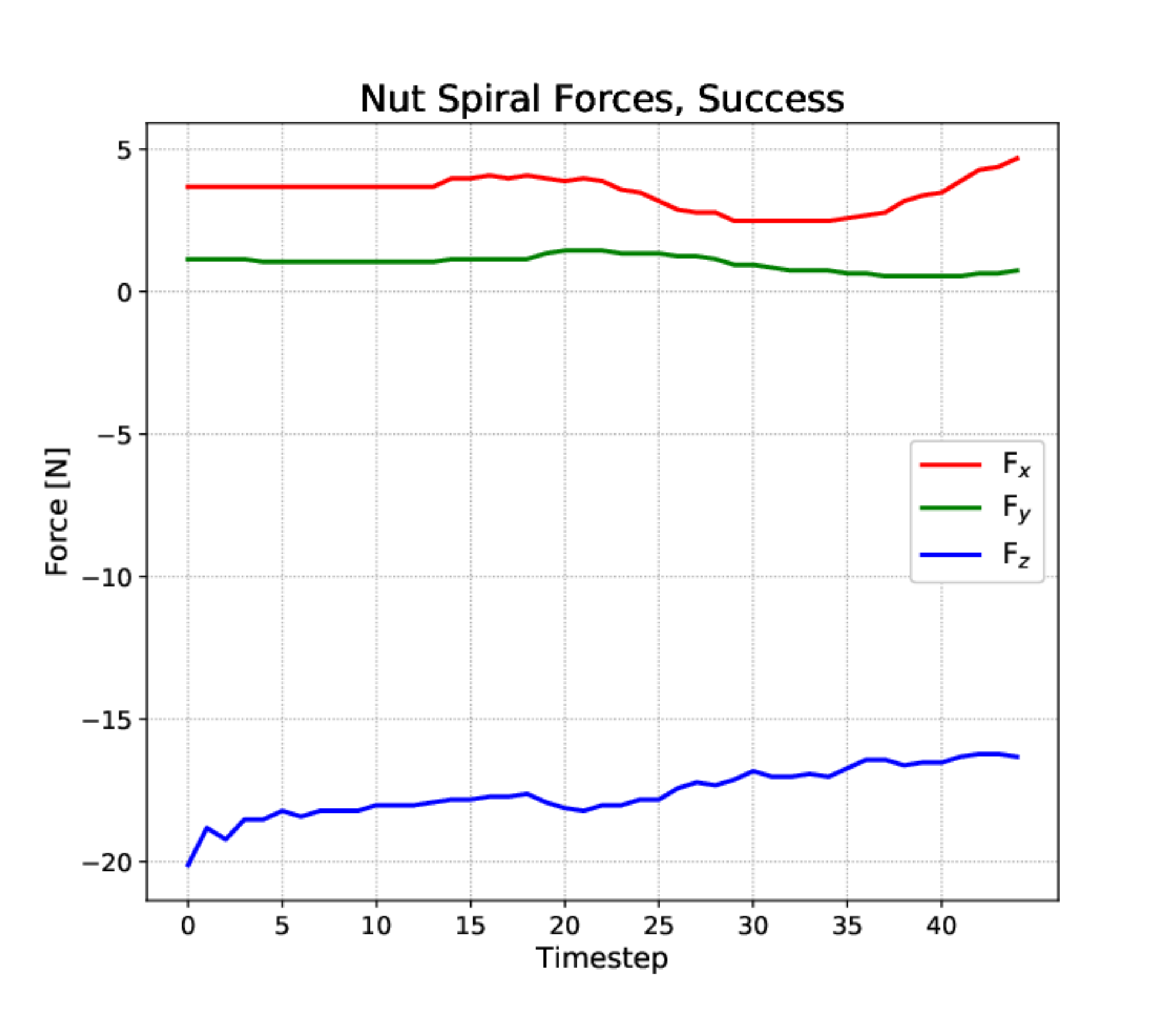}
\includegraphics[height=1.6in]{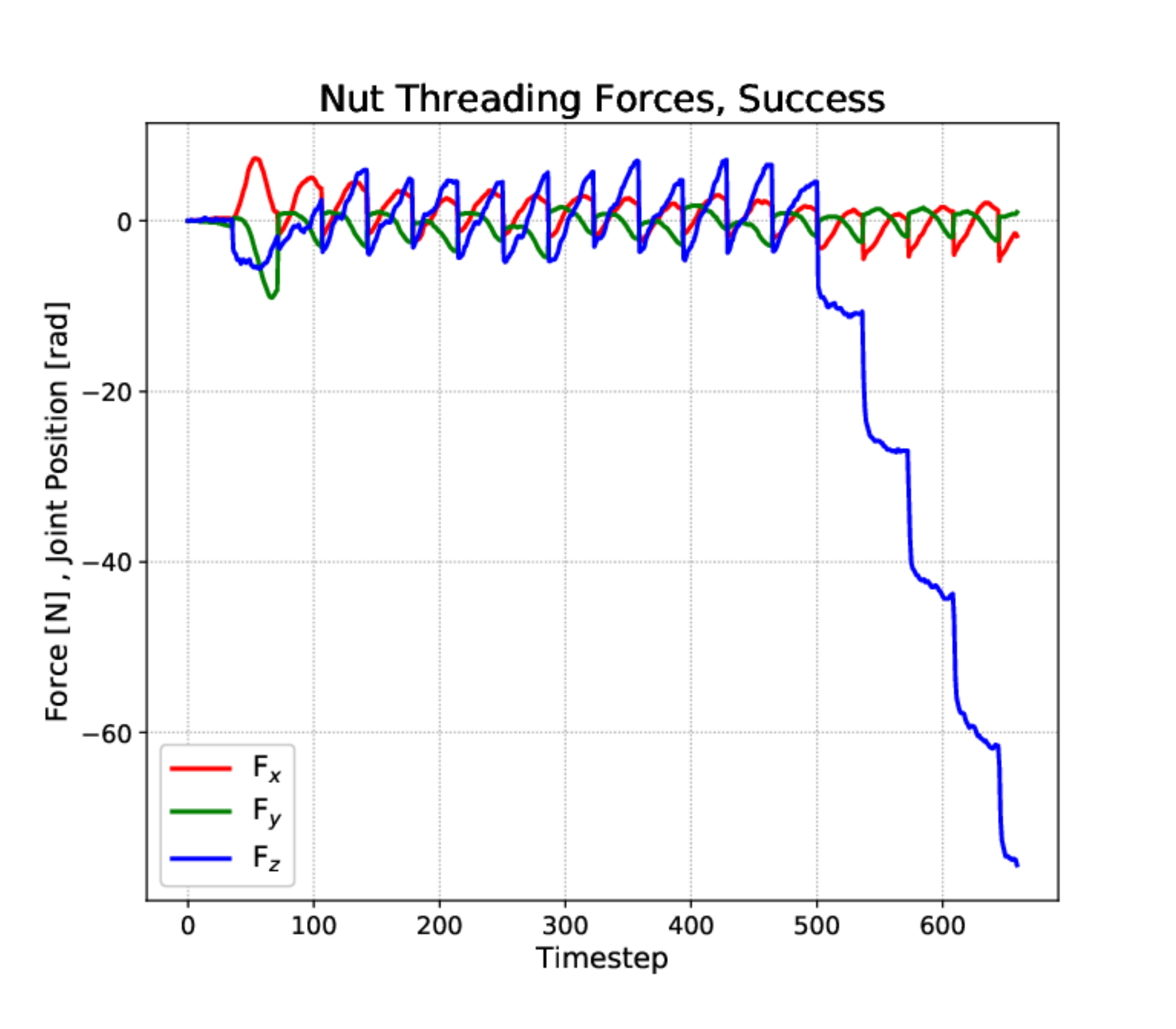}
\includegraphics[height=1.6in]{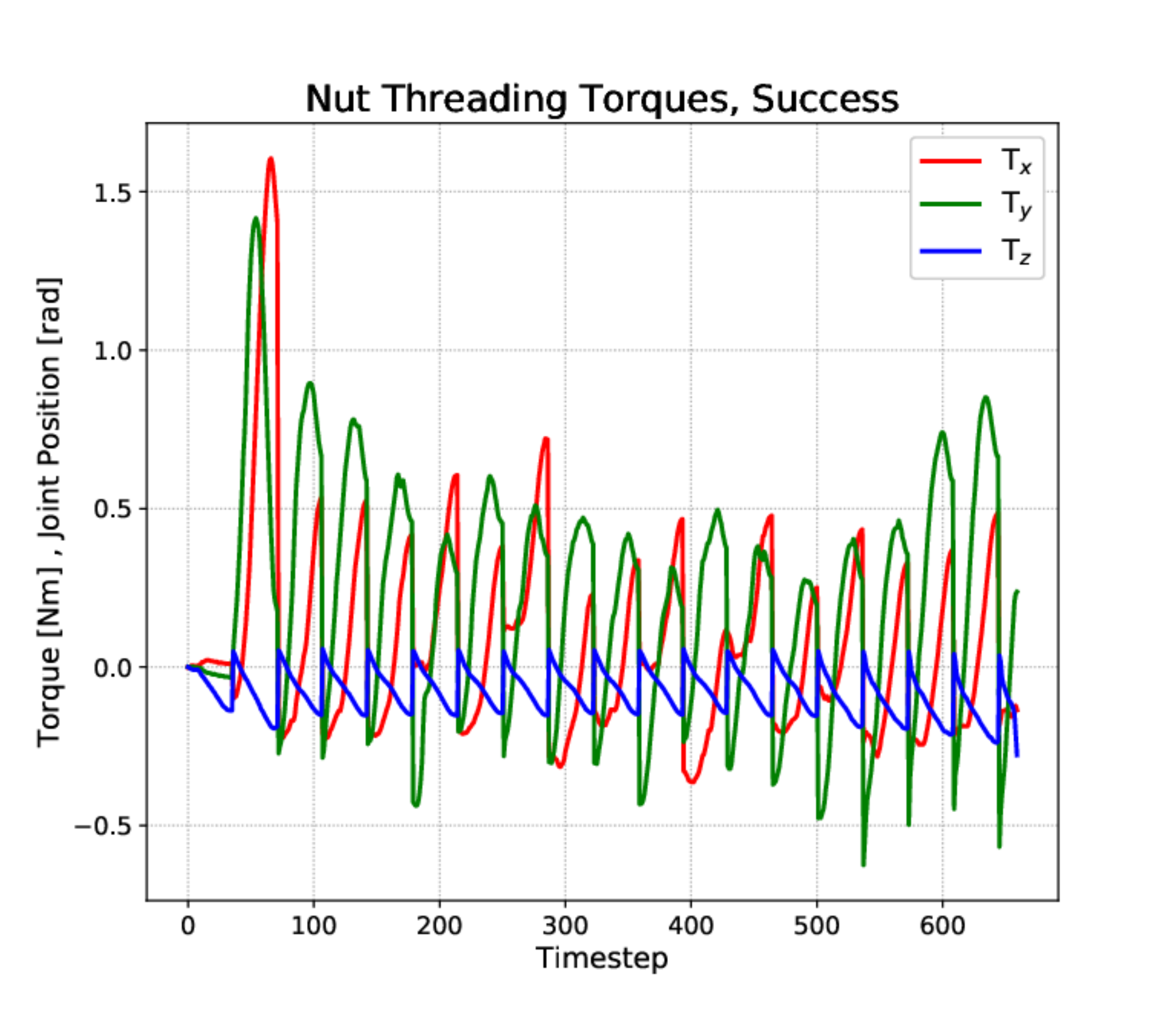}\\
\includegraphics[height=1.6in]{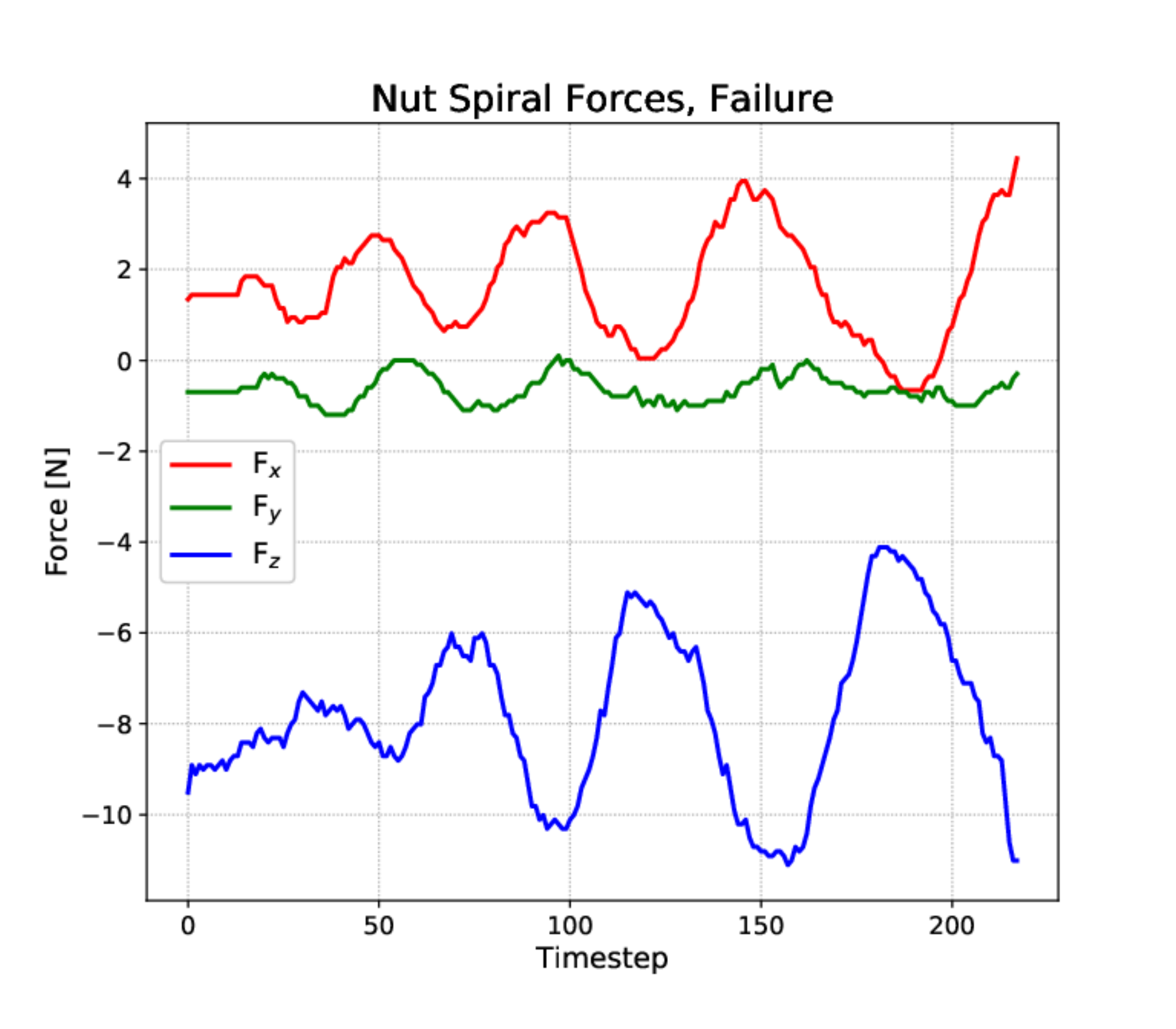}
\includegraphics[height=1.6in]{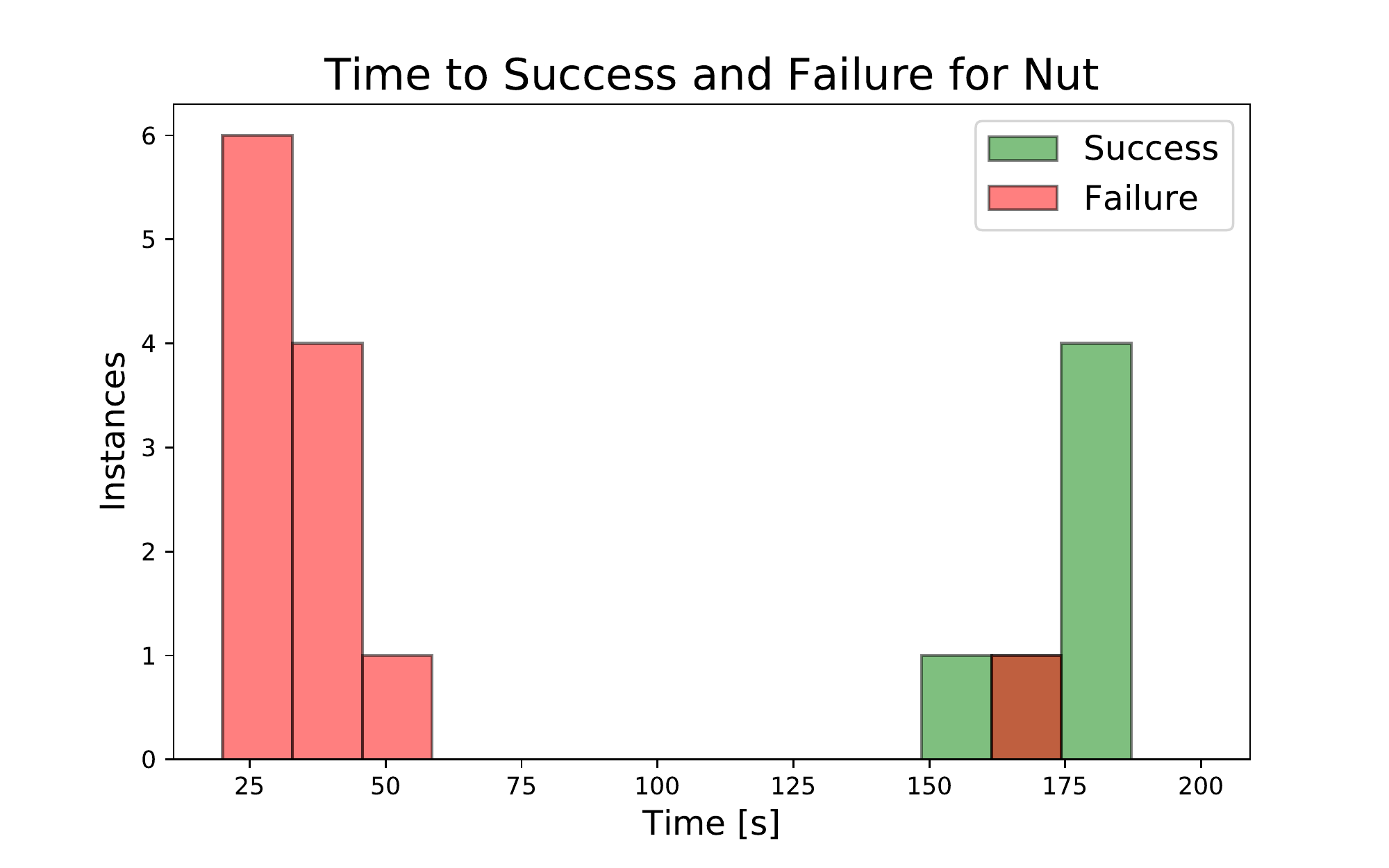} 
\caption{Sample forces and torques during successful (top row) and failed (bottom row) nut assembly, consisting of spiraling (left) and threading motions (center, forces, right, torques). The time distribution of successful (green) and failed (red) attempts are shown to the bottom right.}  
\label{NutTHreadClose}
\label{nutSpiralForceSuccess}
\label{nutThreadForceSuccess}
\label{nutThreadTorqueSuccess}
\label{nutSpiralForceFail}
\end{figure*}

The beginning of the stud approach is also a spiral insertion, but the shape of the stud makes the problem slightly different in ways both advantageous and disadvantageous for the technique applied. The stud (reference part) has a sharp, raised lip on its upper surface that causes friction between it and the nut.  This can cause the nut to stick during points along the spiral action.  In order to overcome this, the stop force criterion had to be raised.  The high friction condition at the top and between the threads also makes it more difficult for the spiral action to recover from the misalignment condition described in Section \ref{pulleyFails}.  This misalignment was the primary reason for task failure.  The spiral insertion action ends with a downward push.  In cases where the action ends with the stud and nut mostly aligned, this had the effect of aligning the nut on the stud, due to the slight taper on the stud and the threaded hole on nut.  This may have been the reason that no cross-threading events were observed during trials.  It may be beneficial for the start of a threaded fastener operation to have some compliance, and this is worth investigating further.

Owing to the shape of the stud and the nut, the majority of task successes occur when the nut is brought immediately to the stud with little to no offset between the stud and nut central axes.  When this occurs, the nut is already mostly ``seated'' on the stud, and the spiral action exits very soon after it begins because the nut has little lateral freedom in this case.  The stopping force criterion is met immediately.  The force profile of a failed spiral operation on the nut looks similar to that of the pulley task, described in Section \ref{pulleyFails}.  The $X$ and $Y$ forces have the expected phase offset, but they have differing zero offsets and magnitudes of oscillation, indicating that the nut has more freedom in the direction that the nut is tilted and hanging off-center from the stud.  In this case the nut falls from the stud as soon as it is released, just as it was for the pulley and the shaft.

If the spiral action successfully places the nut onto the stud, then threading of the nut succeeds in all observed cases.  There were no cross-threading events observed.  The forces seen during the threading action have a saw-tooth profile. The recordings shown in Figures \ref{nutThreadForceSuccess}, top row, include only clockwise motion of the hand, as torque about the stud's axis is applied to the nut. The return motion with the gripper open was not recorded. The most distinguishing feature of the force profile in Figure \ref{nutThreadForceSuccess}, top left, is the $Z$-force at the end of the action, that forms a descending staircase pattern.  It was at this point in the threading operation that the tips of the gripper fingers were in contact with the base plate, and exerting more and more force on the base plate as the hand lowered with each turn.  It was necessary for the gripper to grip the nut quite low, so that as much as the gripper surface as possible was in contact with the flats of the nut. If the gripper happened to slide downwards at all during the finalizing downward push of the spiral action, then the gripper fingers would be at or below the bottom of the nut when threading began.  Occasionally, this condition resulted in friction between the fingers and the base plate to exert a torque sufficient to meet the stopping torque criterion for the threading action.

The difference between the mean time to failure and mean time to completion was greatest for the nut threading task.  Nearly all failures of this task manifested during the spiral action. The threading action takes about two minutes to complete, and most of the observed successful trials completed in about 3 minutes.

\section{Discussion} 
The results show that simple, action-dependent force criteria, such as those that serve as inputs to Algorithms \ref{alg:spiral} and \ref{alg:tilt}, are appropriate to govern common assembly actions.  Although the success rates in Table \ref{successTable} suggests that there is significant room for improvement, it is unclear whether finding better parameters would actually help much, or if we should not rather investigate better methods for early error detection and correction. 

Stopping criteria for our assembly actions have been hand-coded, and the functions we developed take part-specific inputs. Although simple criteria consisting of one to three statements lead to robust results, appropriate criteria could also be replaced by a random forest classifier that is trained on appropriate force-torque data.  We anticipate that a successful classifier would be able to provide rich information about the evolution of assembly actions, and might result in improved assembly performance, albeit at loss of generality. 

We are also interested in online error detection. Take the spiral failures discussed in Sections \ref{pulleyFails} and \ref{studEvolv} as an example.  In this case, the system encountered a situation that the action design did not account for: that the held part was tilted within the hand and hung insecurely on the peg.  However, even though the action is not designed to sense the tilt of the held part directly, the failure mode observed has a distinct force profile that identifies the misalignment.  Identification of this tilted state could trigger a regrasp operation that prepares the system for another spiral insertion attempt.

Although we show only samples here, successful and failed trials show clearly discernible force/torque patterns. In addition, we observe that, albeit overlapping, success and failure have significantly different time distributions. Together, these information might be used to predict whether an ongoing action will be successful or not as a function of time, and possibly even aborted before it fails in a way that is harder to recover from. For example, a failure in placing the nut will lead to the nut falling onto the plate where it might be hard to grasp. If we can detect this failure before releasing the nut, it could be placed back into the kitting tray, and the attempt could be restarted. 

Using touch as a stopping criterion when approaching a part, sometimes fails, for example when the part moves before contact is made. In this case, the robot moves until $\Delta_{max}$ is reached. A simple improvement would therefore be to catch this event and restart the part localization algorithm.  If the system were trained with the force profiles of the various failure modes, then situation-appropriate compensatory actions can be associated with those modes.  Such would form the basis for a robust and flexible task-level planner.

In Section \ref{sct:results}, we discuss how the forces and torques felt at the wrist over the course of action execution correspond to the physical interactions between the held part and the reference part.  The results shown in the figures are intuitively explainable given the relative motion of the parts and the forces we would expect them to impart on each other.  In order to have a system that can properly account for failures encountered while installing a part, it is necessary to have an estimate of the present state of the part.  With a part pose estimate and a model of the part, we can relate the force-torque state of the held object to the geometry of the problem.  This is the fundamental missing piece in the system presented, even if it were equipped with an extensive failure classifier.  

Rosman et al. \cite{sensorPlacement} have developed a method for predicting the best sensor locations to reduce uncertainty in part poses.  It may be possible to use a similar method to plan ``sensing motions'' that would yield the most information about the relationships between parts given models of the effector sensors. 

\section{Conclusion} 

We present a system consisting of a smart gripper with integrated 3D perception and force/torque sensing ability as well as a suite of algorithms to perform standard assemblies. The system can be easily configured to accomplish common assembly tasks.  We have shown that the perceptual modes available to the system are sufficient to accomplish common assembly tasks.  The force and torque data indicate failure modes that are caused by explainable physical interactions between parts in the assembly problem and, we hope to automatically identify these phenomena in future work.  We establish a baseline success rate for our system to accomplish these common tasks, and propose future avenues of investigation that will enable us to improve on that baseline.



\bibliographystyle{tfnlm}
\bibliography{IDSuccess.bib} 

\begin{thebibliography}{10}
\providecommand{\url}[1]{\normalfont{#1}}
\providecommand{\urlprefix}{Available from: }

\bibitem{FDICNN}
Moreira~GR, Lahr~GJ, Boaventura~T, et~al. Online prediction of threading task
  failure using convolutional neural networks. In: 2018 IEEE/RSJ International
  Conference on Intelligent Robots and Systems (IROS); IEEE; 2018. p.
  2056--2061.

\bibitem{dorsey2012efficient}
Dorsey~J, Doggett~W, Hafley~R, et~al. An efficient and versatile means for
  assembling and manufacturing systems in space. In: AIAA SPACE 2012 Conference
  \& Exposition; 2012. p. 5115.

\bibitem{belvin2016space}
Belvin~WK, Doggett~WR, Watson~JJ, et~al. In-space structural assembly:
  Applications and technology. In: 3rd AIAA Spacecraft Structures Conference;
  2016. p. 2163.

\bibitem{komendera2015precise}
Komendera~E, Correll~N. Precise assembly of 3d truss structures using mle-based
  error prediction and correction. The International Journal of Robotics
  Research. 2015;\hspace{0pt}34(13):1622--1644.

\bibitem{prototypeCompose}
N{\"a}gele~F, Halt~L, Tenbrock~P, et~al. A prototype-based skill model for
  specifying robotic assembly tasks. In: 2018 IEEE International Conference on
  Robotics and Automation (ICRA); IEEE; 2018. p. 558--565.

\bibitem{asmPrimitives}
Halt~L, Nagele~F, Tenbrock~P, et~al. Intuitive constraint-based robot
  programming for robotic assembly tasks. In: 2018 IEEE International
  Conference on Robotics and Automation (ICRA); IEEE; 2018. p. 520--526.

\bibitem{iTASC}
Decr{\'e}~W, Smits~R, Bruyninckx~H, et~al. Extending itasc to support
  inequality constraints and non-instantaneous task specification. In: 2009
  IEEE International Conference on Robotics and Automation; IEEE; 2009. p.
  964--971.

\bibitem{lopes1998feature}
Lopes~LS, Camarinha-Matos~LM. Feature transformation strategies for a robot
  learning problem. In: Feature extraction, construction and selection.
  Springer; 1998. p. 375--391.

\bibitem{camarinha1996integration}
Camarinha-Matos~LM, Lopes~LS, Barata~J. Integration and learning in supervision
  of flexible assembly systems. IEEE Transactions on Robotics and Automation.
  1996;\hspace{0pt}12(2):202--219.

\bibitem{forceTraceSVM}
Rodriguez~A, Bourne~D, Mason~M, et~al. Failure detection in assembly: Force
  signature analysis. In: 2010 IEEE International Conference on Automation
  Science and Engineering; IEEE; 2010. p. 210--215.

\bibitem{FSMclassify}
Majdzik~P, Akielaszek-Witczak~A, Seybold~L, et~al. A fault-tolerant approach to
  the control of a battery assembly system. Control Engineering Practice.
  2016;\hspace{0pt}55:139--148.

\bibitem{collabSafety}
Guiochet~J, Machin~M, Waeselynck~H. Safety-critical advanced robots: A survey.
  Robotics and Autonomous Systems. 2017;\hspace{0pt}94:43--52.

\bibitem{corvidAssembly}
von Bayern~AMP, Danel~S, Auersperg~A, et~al. Compound tool construction by new
  caledonian crows. Scientific reports. 2018;\hspace{0pt}8(1):15676.

\bibitem{cobot}
Colgate~JE, Edward~J, Peshkin~MA, et~al. Cobots: Robots for collaboration with
  human operators. 1996;\hspace{0pt}.

\bibitem{UR5manual}
Robots~U. User manual: Ur5 ; 2018.

\bibitem{IndustryCollabIncrease}
Villani~V, Pini~F, Leali~F, et~al. Survey on human--robot collaboration in
  industrial settings: Safety, intuitive interfaces and applications.
  Mechatronics. 2018;\hspace{0pt}55:248--266.

\bibitem{EduCollabIncrease}
Fast-Berglund~{\AA}, Palmkvist~F, Nyqvist~P, et~al. Evaluating cobots for final
  assembly. Procedia CIRP. 2016;\hspace{0pt}44:175--180.

\bibitem{houghCircle}
Hough~PV. Method and means for recognizing complex patterns ; 1962. US Patent
  3,069,654.

\bibitem{ICP}
Chen~Y, Medioni~G. Object modelling by registration of multiple range images.
  Image and vision computing. 1992;\hspace{0pt}10(3):145--155.

\bibitem{van2018comparative}
Van~Wyk~K, Culleton~M, Falco~J, et~al. Comparative peg-in-hole testing of a
  force-based manipulation controlled robotic hand. IEEE Transactions on
  Robotics. 2018;\hspace{0pt}34(2):542--549.

\bibitem{sensorPlacement}
Rosman~G, Choi~C, Dogar~M, et~al. Task-specific sensor planning for robotic
  assembly tasks. In: 2018 IEEE International Conference on Robotics and
  Automation (ICRA); IEEE; 2018. p. 2932--2939.

\end{thebibliography}

\end{document}